\documentclass[10pt,twocolumn,letterpaper]{article}

\usepackage[pagenumbers]{cvpr} 

\definecolor{cvprblue}{rgb}{0.21,0.49,0.74}
\usepackage[pagebackref,breaklinks,colorlinks,allcolors=cvprblue]{hyperref}
\usepackage{multirow}
\usepackage{xcolor}
\usepackage{colortbl, booktabs}
\usepackage{array}
\usepackage{enumitem}
\usepackage{graphicx}
\usepackage{longtable}
\usepackage{booktabs}
\usepackage{subcaption}
\usepackage{parskip}
\usepackage{algorithm}  
\usepackage{algorithmic}
\usepackage{amsmath}
\usepackage{relsize}

\newcommand{\algcomment}[1]{
  $\triangleright$ \textit{#1} 
}

\def\paperID{4204} 
\def\confName{CVPR}
\def\confYear{2026}

\title{GeoTikzBridge: Advancing Multimodal Code Generation \\ for Geometric Perception and Reasoning}

\author{Jiayin Sun, Caixia Sun, Boyu Yang$^*$, Hailin Li, Xiao Chen, Yi Zhang\\ Errui Ding, Liang Li, Chao Deng, Junlan Feng
\\
JIUTIAN Research\\
{\tt\small \{sunjiayin, yangboyu, liliang\}@cmjt.chinamobile.com} \\
}

\begin{document}
\maketitle
\begin{abstract}





%

Multimodal Large Language Models (MLLMs) have recently demonstrated remarkable perceptual and reasoning abilities.
However, they struggle to perceive fine-grained geometric structures, constraining their ability of geometric understanding and visual reasoning.
To address this, we propose GeoTikzBridge, a framework that enhances local geometric perception and visual reasoning through tikz-based code generation.
Within this framework, we build two models supported by two complementary datasets.
The GeoTikzBridge-Base model is trained on GeoTikz-Base dataset, the largest image-to-tikz dataset to date with 2.5M pairs (16 $\times$ larger than existing open-sourced datasets).
This process is achieved via iterative data expansion and a localized geometric transformation strategy.
Subsequently, GeoTikzBridge-Instruct is fine-tuned on GeoTikz-Instruct dataset which is the first instruction-augmented tikz dataset supporting visual reasoning.
Extensive experimental results demonstrate that our models achieve state-of-the-art performance among open-sourced MLLMs.
Furthermore, GeoTikzBridge models can serve as plug-and-play reasoning modules for any MLLM(LLM), enhancing reasoning performance in geometric problem-solving. Datasets and codes are publicly available at: \textcolor{blue}{\href{https://github.com/sjy-1995/GeoTikzBridge}{Link}}.
{\color{red}
%
%
%
%
}

\end{abstract}    
\footnotetext{$^*$Corresponding author.}
\section{Introduction}
\label{sec:intro}


Multimodal Large Language Models (MLLMs) have made remarkable progress in cross-modal perception and reasoning, demonstrating capabilities covering from fine-grained visual understanding to complex mathematical problem scenarios. 
In contrast, geometric problem remains challenging, as it demands the integration of fine-grained geometric perception and structured symbolic reasoning. This capability is fundamental for high-level cognitive and scientific tasks such as diagram-based mathematics, spatial reasoning and STEM diagram interpretation. Bridging this gap is crucial for advancing MLLMs toward deeper geometric cognition and interpretable reasoning.

Recent studies have explored solutions from the perspective of image-to-code generation and geometric reasoning. Existing Img2Code methods usually focus on web UI-to-HTML/CSS or data charts-to-Python\cite{wu2025webdancer, jiang2025screencoder, beltramelli2018pix2code, roberts2024image2struct, kaluarachchi2024webdraw, wu2024plot2code, zhao2025chartcoder, xing2025chartcode, yen2025code}. Though a few works have touched on reconstructible code generation for graphics, they rarely involve geometric contents\cite{wang2025mathcoder, belouadi2025tikzero, rodriguez2025starvector, yang2025omnisvg}. For the downstream reasoning task, existing mathematical reasoning models primarily rely on text-based reasoning\cite{Qwen2.5-VL, qwen3technicalreport, zhu2025internvl3, wang2025internvl3_5, vteam2025glm45vglm41vthinkingversatilemultimodal}, while often overlooking the geometric relational transfer required for visual reasoning in geometric problem-solving.

\begin{figure*}[t]
  \centering
    \includegraphics[width=16.8cm, height=7.8cm]{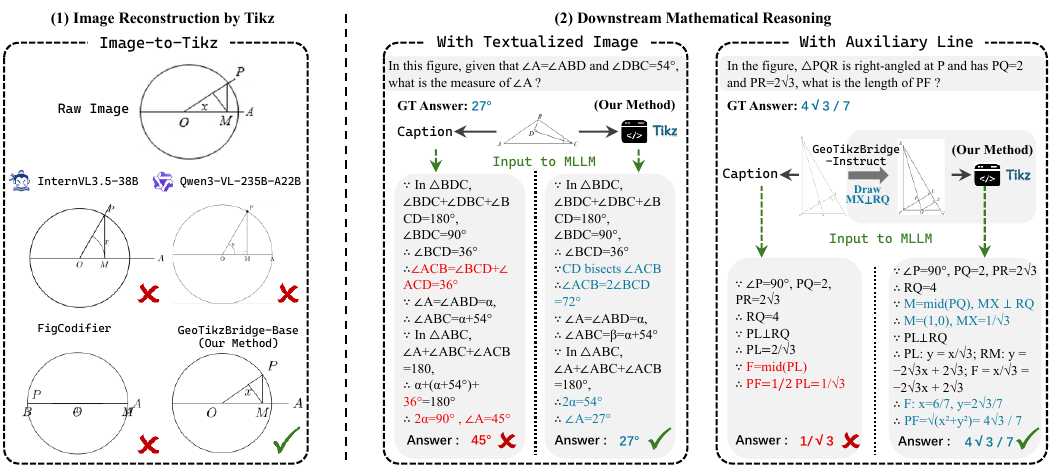}
    \vspace{-8pt}
    \caption{GeoTikzBridge demonstrates advantages in geometric perception and mathematical reasoning. (1) GeoTikzBridge-Base achieves the most accurate reconstruction of local geometric structures compared to existing approaches. (2) The generated tikz representations enhance MLLMs’ mathematical and visual geometric reasoning.}\vspace{-10pt}
    \label{fig: instruction_fig1}
\end{figure*}

Despite these advancements, MLLMs still exhibit limited local geometric perception, struggling to parse fine-grained visual details such as segment relations, angle magnitudes, and shape constraints (Fig.~\ref{fig: instruction_fig1}(1)).
This limitation is primarily due to the scarcity of large-scale geometric image–code datasets and insufficient modeling of subtle geometric variations.

To address these challenges, we propose GeoTikzBridge, a framework designed for enhancing multimodal geometric perception via structured tikz code.
To alleviate data scarcity, we first construct the GeoTikz-Base dataset, the largest image-to-tikz dataset containing 2.5M image–tikz pairs, by iteratively expanding a high-quality seed dataset.
To further enhance the model perception on geometric details, we propose a localized geometric transformation strategy, including a code transformation method that leverages localized code editing to enhance geometric details in codes.
Building upon this strategy, we additionally develop GeoTikz-Instruct, an auxiliary dataset that enables models to generate and interpret instruction-guided auxiliary lines.

Notably, the generated tikz codes can be seamlessly integrated into downstream geometric reasoning tasks, complementing MLLMs' geometric perception capabilities through structured textual encoding of geometric elements and relationships, as illustrated in Fig.~\ref{fig: instruction_fig1}(2).

The main contributions of this paper can be summarized as follows:

\begin{itemize}
    \item We propose GeoTikzBridge, a novel framework that bridges geometric perception and symbolic reasoning via tikz-based code generation, thereby enabling visual reasoning in MLLMs.
    \item We construct two datasets: GeoTikz-Base, the largest image-to-tikz dataset to date and GeoTikz-Instruct, the first instruction-augmented tikz dataset for visual reasoning.
    \item We develop GeoTikzBridge models which achieve state-of-the-art performance among open-sourced MLLMs and can serve as plug-and-play reasoning modules to enhance geometric reasoning capabilities.
\end{itemize}

\begin{figure*}[t]
  \centering
    \includegraphics[width=16.8cm, height=7.8cm]{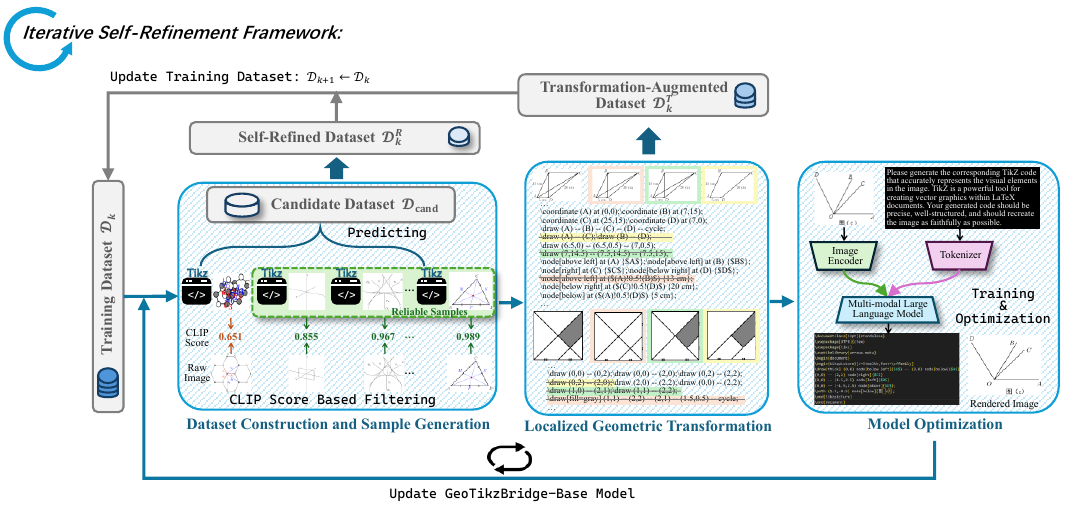}
    \vspace{-8pt}
    \caption{Framework of the proposed iterative self-refinement for GeoTikz-Base dataset and GeoTikzBridge-Base model. It iteratively conducts three steps: dataset construction and sample generation, localized geometric transformation, and model optimization.}\vspace{-10pt}
    \label{fig: overview}
\end{figure*}


\section{Related Works}
\label{sec:related}

\textbf{Image-to-Code Generation.} Image-to-Code Generation allows multimodal systems to translate visual content into structured symbolic forms. These forms can be executed or further edited. Recent studies have achieved notable progress across various code forms, including vector graphics\cite{rodriguez2025starvector, yang2025omnisvg, cai2023leveraging, wu2023iconshop}, HTML/CSS\cite{wu2025webdancer, jiang2025screencoder, beltramelli2018pix2code, roberts2024image2struct, kaluarachchi2024webdraw}, Python\cite{wu2024plot2code, zhao2025chartcoder, xing2025chartcode, yen2025code}, and other codes\cite{deng2019codeegan, pang2020novel}. 
However, most of these efforts focus on converting webpage or chart images into executable code, leaving critical gaps in geometric code generation. Only a few works\cite{wang2025mathcoder, belouadi2025tikzero, rodriguez2025starvector, yang2025omnisvg} have explored converting more diverse graphic images into codes that reproduce the visual elements. Among these, tikz\cite{tikz_url},
a LaTeX-based graphic language, is superior to formats like SVG for geometric reasoning because its structured, code-based syntax inherently documents the logical steps and dependencies of geometric constructions, not just their final visual appearance.
DaTikZ\cite{belouadi2025tikzero} pioneered pairing graphic images with tikz codes, enabling precise coding of elements and their spatial relationships. However, its geometric samples remain limited, focusing mainly on general graphics rather than the precise line segments, angles, and shapes needed for geometric tasks.  
To address this gap, we propose GeoTikz-Base, a large-scale geometric image-to-tikz dataset to alleviate the scarcity of high-quality geometric-dominant image-tikz data.


\textbf{Multimodal Mathematical Reasoning.} 
Mathematical reasoning has become an important capability of MLLMs, yet geometric problem-solving remains challenging due to the need for fine-grained visual perception.
Existing mathematical reasoning works could be roughly divided into generalized and specialized methods.
Generalized MLLMs\cite{Qwen2.5-VL, qwen3technicalreport, zhu2025internvl3, wang2025internvl3_5, vteam2025glm45vglm41vthinkingversatilemultimodal} exhibit strong cross-modal performance, however, they are limited by scarce high-quality geometric visual-code data and poor sensitivity to local structures, which constrain their geometric-specific performance\cite{zhao2025towards, mouselinos2024beyond}.
Specialized methods focus on geometry-specific optimizations\cite{yang2024mathglm, guo2024mammoth, li2024eagle, sun2025mathglance, pan2025enhancing, yang2025bridging, chen2025geopqa, zhao2025towards} through improved visual-text alignment, fine-tuning on fine-grained datasets, symbolic based intermediate reasoning, etc. However, they lack local geometric perception and auxiliary-line reasoning, limiting their effectiveness in geometric tasks.
To address this, we develop GeoTikzBridge-Base and GeoTikzBridge-Instruct models based on the constructed image-to-tikz datasets, which could boost geometric-dominant reasoning by the generated tikz codes and auxiliary lines.



\section{Method}
\label{sec:method}

\subsection{Overview}
We present the GeoTikzBridge framework, which advances visual reasoning by bridging the gap between geometric images and structured representations, specifically through the generation of tikz code. The GeoTikzBridge framework consists of three components: (i) an iterative self-refinement framework for constructing the image-to-tikz dataset called GeoTikz-Base and training the corresponding model called GeoTikzBridge-Base, (ii) an instruction guided image-to-tikz dataset called GeoTikz-Instruct and the corresponding trained model called GeoTikzBridge-Instruct, and (iii) a training-free visual reasoning pipeline in geometric problem-solving.

Specifically, the self-refinement framework begins with a high-quality seed dataset and iteratively expands the training data by generating new image-tikz pairs. In each iteration, tikz code is predicted by the model and unreliable samples are filtered by comparing the original image with its rendering. A novel localized code transformation strategy is introduced to enhance the model sensitivity to fine-grained geometric details. Building upon this foundation, we construct a specialized dataset and train a model for generating auxiliary lines, making it easier to visual reasoning on mathematical problems. Finally, we explore the broader applicability of our approach by proposing a training-free plug-and-play pipeline for visual reasoning.

\subsection{Iterative Self-Refinement for Image-to-Tikz Dataset Construction and Model Training} \label{sec. 3.1}

The scarcity of high-quality image-tikz code pairs for model training necessitates the adoption of a self-refinement strategy. This approach iteratively generates reliable data and retrains the image-to-tikz model. The reliability of the data is assessed by comparing the rendered image with the original image. 

During the iterative self-refinement process, both the training data and the model are continuously updated. For the $k$-th iteration ($k>1$), the updated training dataset $\mathcal{D}_k$ is formed by combining the dataset from the previous iteration $\mathcal{D}_{k-1}$ with the two newly acquired datasets: the self-refined set $\mathcal{D}_{k-1}^R$ and the transformation-augmented set $\mathcal{D}_{k-1}^T$. Here, $\mathcal{D}_k=\{(I_k,C_k)\}$, where $I_k$ and $C_k$ denote an image and the corresponding tikz code in the updated training dataset at the $k$-th iteration, respectively. The model $M_k$ is optimized on the training dataset $\mathcal{D}_k$ under the maximum likelihood objective $\mathcal{L}_\text{gen}$. The iterative process can be formalized as follows:
\begin{align}
\begin{cases}
\mathcal{D}_k = \mathcal{D}_{k-1} \cup \mathcal{D}_{k-1}^R \cup \mathcal{D}_{k-1}^T, \\
M_k = \mathop{\arg\min}\limits_{M \gets M_{k-1}} \mathbb{E}_{(I,C) \sim \mathcal{D}_k} \left[ \mathcal{L}_{\text{gen}}(M(I), C) \right]
\end{cases}
\end{align}
where $k = 1, 2, \ldots, K$, and $K$ denotes the maximum iteration number, which is set to $4$ in our experiments for a balance between cost and performance.
The overall process for image-to-tikz data construction and model training consists of the following three steps:

\textbf{Step1. Dataset Construction and Sample Generation}

As the only open-sourced and large-scale image-tikz code pairing dataset sourced from public arXiv papers and TeX repositories, DaTikZ\cite{belouadi2025tikzero} serves as the seed dataset (denoted as $\mathcal{D}_0$). The model initially trained on its training set with 145k samples is referred to as $M_0$.

Due to the limited variety of geometric images in DaTikZ, which cannot effectively cover diverse geometric scenarios. The goal of this step is to expand the seed dataset with reliable generated samples from a large collection of candidate geometric images.

\textbf{Dataset Construction} We construct a candidate dataset $\mathcal{D}_{\text {cand}}$ by collecting and organizing geometric images from nine public datasets. This candidate dataset comprises a rich collection of unannotated tikz geometric diagrams from diverse sources. The datasets utilized are as follows: MMathCoT-1M\cite{luo2025dataset1}, GEOS\cite{seo2015dataset2}, MAVIS-Instruct-Geometry\cite{zhang2024mavisdatasets3and4}, MAVIS-Instruct-Function\cite{zhang2024mavisdatasets3and4}, geo170k-alignment\cite{datasets5and6}, geo170k-qa-tuning\cite{datasets5and6}, Geometry3K\cite{lu2021interdataset7}, GeomVerse\cite{kazemi2023geomversedataset8}, UniGeo\cite{chen2022unigeodataset9}. These datasets contain both synthetic and real-world images, with a primary focus on basic geometric elements, spatial relationships, and topological structures.

\textbf{Sample Generation.} Guided by the self-refinement paradigm, we select reliable samples from the constructed candidate dataset $\mathcal{D}_{\text {cand}}$ as potential additions to the training set by applying a CLIP score-based filtering strategy. The process begins with the generation of tikz pseudo-label set $\{\hat{C}\}$ for the geometric images, which are predicted by the model $M_{k}$. Then, the CLIP score (a normalized cosine similarity metric) $s$ between the image $\hat{I}$ rendered from the predicted tikz code $\hat{C}$ and the original image $I$ is computed. If $s$ exceeds a threshold $\tau$, the code $\hat{C}$ and the rendered image $\hat{I}$ would be considered as a reliable sample and incorporated in the self-refined set $\mathcal{D}_k^R$. $\mathcal{D}_k^R$ is defined as:
\begin{small} 
\begin{align}
&\mathcal{D}_{k}^{R} = \left\{(\hat{I}, \hat{C}) \mid \hat{C}=M_k(I), \hat{I}=\mathcal{R}(\hat{C}), s(I,\hat{I})>\tau\right\}
\end{align}
\end{small} 
where $\mathcal{R}(\cdot)$ denotes the render operation and $I \in \mathcal{D}_{\text {cand}}$.

\textbf{Step2. Localized Transformation Strategy}

\begin{figure}[t]
  \centering
  \includegraphics[width=6.4cm, height=5.0cm]{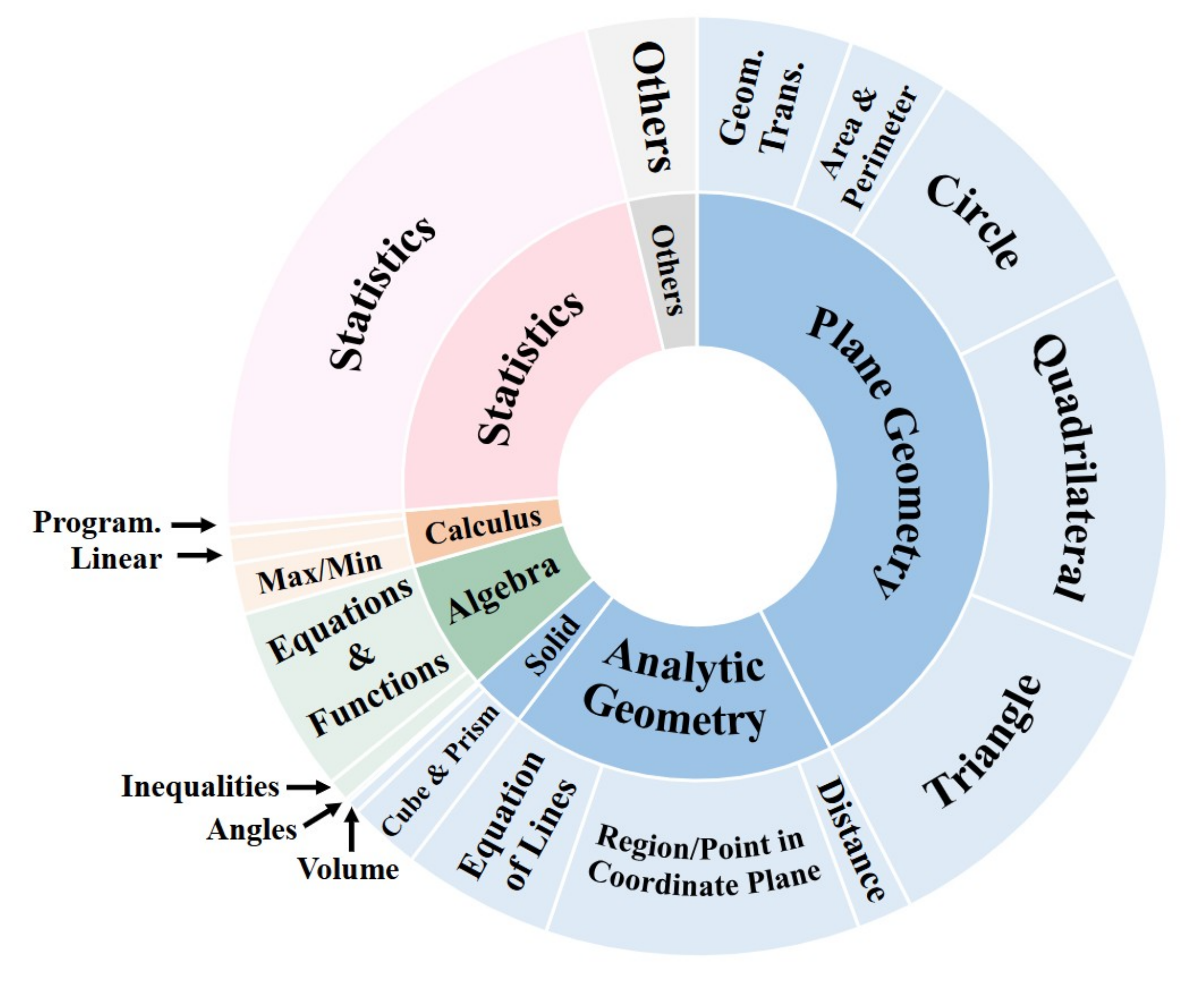}
    \caption{Distribution of the 2.5M GeoTikz-Base dataset.}
    \label{fig: base_dataset_distribution}
\end{figure}

We observed that complex images often cause the model to neglect fine-grained geometric details, resulting in the omission or hallucination of critical code lines. To address this, we propose a localized transformation strategy applied to the tikz codes and their corresponding images to better capture geometric features.

The transformation strategy consists of two components: code transformation applied during the iterative dataset expansion, and image transformation applied during the model optimization (see Supplementary Material for details). Code transformation is operated on each tikz code in the self-refined set $\mathcal{D}_k^R$ at the $k$-th iteration, which randomly removes $1$ to $n$ lines (where $n$ does not exceed 40\% of the total tikz code lines splitted with the semicolon). Each compilable modified code $\tilde{C}$ along with its rendered image $\tilde{I}$ serve as a sample pair in the transformation-based incremental dataset $\mathcal{D}_k^T$ at the $k$-th iteration.
Denote the transformation operation as $\mathcal{T}(\cdot)$, $\mathcal{D}_k^T$ can be defined as:
\begin{small} 
\begin{align}
&\mathcal{D}_k^T=\left\{(\widetilde{I},\widetilde{C}) \mid \widetilde{I}=\mathcal{R}(\widetilde{C}), \widetilde{C}=\mathcal{T}(\hat{C}), \hat{C}\in\mathcal{D}_k^R \right\}
\end{align}
\end{small}

Through code transformation, the model is forced to learn the structural and fine-grained semantics of the code rather than memorizing specific text sequences. It can also be regarded as injecting code noise, which enhances the generalization and robustness of the code generation. Leveraging this strategy, the code repetition prediction rate decreases by 15\%. The finally expanded 2.5M training samples is called GeoTikz-Base. Fig. \ref{fig: base_dataset_distribution} shows that the distribution of GeoTikz-Base predominantly features Plane Geometry and Analytic Geometry as core components, while also incorporating Algebra, Calculus, and other mathematical domains, exhibiting comprehensive coverage of geometric figures.

\textbf{Step3. Model Optimization}

For the $k$-th iteration, the model $M_{k}$ is initialized from the previous model $M_{k-1}$ and optimized on the $k$-th training dataset $\mathcal{D}_k$. The training follows the causal auto-regressive modeling paradigm, where each code token in the tikz code sequence $c_i \in C$ is predicted conditioned on the image $I$ and all previous tokens $c_{<i}$. The corresponding training objective is defined as:
\begin{align}
\mathcal{L}_{\mathrm{gen\ }}\left(M\left(I\right),C\right)=-\sum_{i}{\log{\mathrm{P}_{M}}\left(c_i\mid I,c_{<i}\right)},
\end{align}
where $\mathrm{P}_{M}$ denotes the probability prediction by model $M$.

\subsection{Instruction Guided Image-to-Tikz Dataset Construction and Model Training}

To address the challenge of auxiliary line generation in geometric problems, we not only construct the GeoTikz-Instruct dataset \(\mathcal{D}_{\text{ins}}\) based on the built $\mathcal{D}_K$, but also train the auxiliary line generation model GeoTikzBridge-Instruct $M_{\text{ins}}$ on \(\mathcal{D}_{\text{ins}}\).

Firstly, we apply the code transformation strategy \(\mathcal{T}(\cdot)\) (introduced in Sec. \ref{sec. 3.1}) to each sample in \(\mathcal{D}_K\), yielding the transformed tikz code \(\tilde{C}_K\) and its rendered image \(\tilde{I}_K\). Then, these images are fed into Qwen2.5-VL-72B to annotate the instruction $Q$ describing what auxiliary lines have been added or what geometric elements have been modified. Finally, to ensure dataset quality, we use Doubao to filter unreliable annotations via the VLM-based filter strategy \(\mathcal{F}(\cdot)\). Each retained annotated instruction $Q'$ along with its corresponding pre-transformed tikz code $C'$ and post-transformed image $\tilde{I}'$ (rendered from post-transformed tikz code $\tilde{C}'$) form a sample pair in the GeoTikz-Instruct dataset \(\mathcal{D}_{\text{ins}}\). \(\mathcal{D}_{\text{ins}}\) is defined as follows:
\begin{small} 
\begin{align}
\mathcal{D}_{\text{ins}} = \left\{ (Q', \tilde{I}', C') | \begin{aligned}
&\tilde{I}' = \mathcal{F}(\tilde{I}_K), \tilde{I}_K=\mathcal{R}(\mathcal{T}(C_K)),  \\
&Q'= \mathcal{F}(Q), C' = \mathcal{F}(C_K)
\end{aligned}\right\}
\end{align}
\end{small} 
where $(I_K,C_K) \in \mathcal{D}_K$. The built GeoTikz-Instruct dataset contains 419k training samples.

Then, we develop an instruction-guided auxiliary line generation model GeoTikzBridge-Instruct $M_\text{ins}$ based on the built \(\mathcal{D}_{\text{ins}}\) by performing SFT on the GeoTikzBridge-Base model. The optimization is defined as follows:
\begin{small}
\begin{align}
M_{\text{ins}} = \arg\min_M \mathbb{E}_{(Q',\tilde{I}',C') \sim \mathcal{D}_{\text{ins}}} \left[ \mathcal{L}_{\text{gen}}(M(\tilde{I}',Q'), C') \right]
\end{align}
\end{small}

\subsection{Training-Free Visual Reasoning in Geometric Problem-Solving}

VLMs struggle with geometric problems due to two issues: visual feature compression causes critical detail loss, and auxiliary line requirements complicate COT reasoning. Besides, LLMs excel at reasoning but cannot directly process geometric images (e.g., via OCR), limiting their applicability. To address these issues, we design a training-free visual reasoning pipeline integrating GeoTikzBridge-Base or GeoTikzBridge-Instruct as plug-and-play modules for VLMs and LLMs. 

For VLM reasoning with tikz codes, GeoTikzBridge-Base is utilized for translating the geometric problem image $I_p$ to tikz code $C_p$. It is noted that for LLMs, prompts only include problem text $T_p$ and tikz code $C_p$.

For VLM reasoning with auxiliary lines, Algorithm \ref{alg:geo_solving} shows the pipeline, where the VLM itself uses original image $I_p$ and problem text $T_p$ to 
predict the auxiliary line instruction $P_\text{aux}$ (None if auxiliary lines are unnecessary determined by the VLM itself)
via prompt $P_{\text{ins}}$ (see Supplementary Material for details). If no auxiliary lines are needed, the VLM then solves the problem with $T_p$, $I_p$, and prompt for problem solving $P_\text{ps}$ directly; Otherwise, GeoTikzBridge-Instruct generates the tikz code $C_{\text{aux}}$ with auxiliary lines added based on $P_{\text{aux}}$, which is then rendered to $I_{\text{aux}}$. The VLM then utilizes $T_p$, $I_p$, $P_\text{ps}$ together with $I_{\text{aux}}$ and $C_{\text{aux}}$ for problem solving.

\begin{algorithm}[h]
\caption{\small{Geometric Reasoning with Auxiliary Line}}
\label{alg:geo_solving}
\begin{footnotesize}
\algorithmicrequire{ Problem text $ T_p $, problem image $ I_p $, prompt $P_{\text{ins}}$ for generating auxiliary lines, prompt $P_\text{ps}$ for problem solving}, and Vision-Language Model (VLM) \\
\algorithmicensure{ Solution $ A $ to the problem }
\begin{algorithmic}[1]
    \STATE $ P_{\text{aux}} \gets \text{VLM}(I_p, T_p, P_{\text{ins}}) $ \hspace{4.2em}  \algcomment{\scriptsize{Query VLM for auxiliary lines}}
    \IF{$ P_{\text{aux}} \text{ is None} $}  
        \STATE $ A \gets \text{VLM}(I_p, T_p, P_{\text{ps}}) $ \hspace{4.2em}  \algcomment{\scriptsize{Solve without auxiliary lines}}
    \ELSE  
        \STATE $ C_{{\text{aux}}} \gets \text{GeoTikzBridge-Instruct}(I_p, P_{\text{aux}}) $ 
        \STATE $ I_{{\text{aux}}} \gets \mathcal{R}(C_{{\text{aux}}}) $
        \STATE $ A \gets \text{VLM}(T_p, I_p, P_{\text{ps}}, I_{{\text{aux}}}, C_{{\text{aux}}}) $ \hspace{-0.2em}  \algcomment{\scriptsize{Solve with auxiliary lines}}
    \ENDIF  \\
\algorithmicreturn{ Solution $A$}
\end{algorithmic}
\end{footnotesize} 
\end{algorithm}

\begin{table*}[t]
\renewcommand{\arraystretch}{0.8}
\footnotesize
  \caption{Performance comparison of image-to-tikz methods across four datasets. The best and second-best results are marked with bold and underlines, respectively.}\vspace{-5pt}
  \label{table: GeoTikzBridge-Base_performance}
  \centering
  \begin{tabular}{p{3.3cm}>{\centering\arraybackslash}p{1.1cm}>{\centering\arraybackslash}p{0.5cm}>{\centering\arraybackslash}p{0.5cm}>{\centering\arraybackslash}p{1.1cm}>{\centering\arraybackslash}p{0.5cm}>{\centering\arraybackslash}p{0.5cm}>{\centering\arraybackslash}p{1.1cm}>{\centering\arraybackslash}p{0.5cm}>{\centering\arraybackslash}p{0.5cm}>{\centering\arraybackslash}p{1.1cm}>{\centering\arraybackslash}p{0.5cm}>{\centering\arraybackslash}p{0.5cm}}
    \toprule
    \multirow{2}{*}{Method} & \multicolumn{3}{c}{DaTikZ} & \multicolumn{3}{c}{MathVista-GPS} & \multicolumn{3}{c}{GAOKAO-MM-Math} & \multicolumn{3}{c}{EDUBenchmark} \\
    \cmidrule(lr){2-13} & CLIP-S$\uparrow$ & FID$\downarrow$ & CSR$\uparrow$ & CLIP-S$\uparrow$ & FID$\downarrow$ & CSR$\uparrow$ & CLIP-S$\uparrow$ & FID$\downarrow$ & CSR$\uparrow$ & CLIP-S$\uparrow$ & FID$\downarrow$ & CSR$\uparrow$ \\
    \midrule
    \rowcolor{gray!30}
    \multicolumn{13}{c}{\textit{Open-sourced VLMs}} \\
    \midrule
    Qwen2.5-VL-32B\cite{Qwen2.5-VL} & 0.788 & 77.2 & 80.6\% & 0.832 & 61.0 & \underline{97.6\%} & 0.808 & 92.2 & 94.4\% & 0.748 & 87.0 & 81.2\%  \\
    Qwen2.5-VL-72B\cite{Qwen2.5-VL} & 0.795 & 49.8 & 81.5\% & 0.858 & 46.9 & \bf 98.1\% & 0.854 & 61.3 & 90.1\% & 0.781 & 60.5 & 81.8\% \\
    Qwen3-VL-30B-A3B\cite{qwen3technicalreport} & 0.672 & 174.4 & 89.1\% & 0.802 & 65.7 & 81.0\% & 0.826 & 107.9 & \underline{95.2\%} & 0.748 & 95.9 & 80.2\% \\
    InternVL3-78B\cite{zhu2025internvl3} & 0.747 & 62.7 & 78.8\% & 0.860 & 47.9 & 92.3\% & 0.871 & 55.4 & 93.7\% & 0.801 & 58.3 & 81.5\% \\
    InternVL3.5-38B\cite{wang2025internvl3_5} & 0.733 & 50.6 & \underline{94.7\%} & 0.786 & 104.0 & 93.8\% & 0.834 & 77.2 & 91.5\% & 0.768 & 75.6 & \underline{83.1\%} \\
    GLM4.5-V-106B-A12B\cite{vteam2025glm45vglm41vthinkingversatilemultimodal} & \underline{0.806} & 50.9 & 93.8\% & 0.769 & 122.7 & 88.9\% & 0.835 & 75.9 & 77.5\% & \underline{0.809} & 61.4 & 62.7\% \\
    \midrule
    \rowcolor{gray!30}
    \multicolumn{13}{c}{\textit{Existing Image-to-Tikz Model}} \\
    \midrule
    FigCodifier-8B\cite{wang2025mathcoder} & 0.785 & 45.8 & 86.7\% & 0.884 & 42.5 & 88.9\% & 0.849 & 59.1 & 92.3\% & 0.675 & 67.9 & 67.3\% \\
    \midrule
    GeoTikzBridge-Base-8B & 0.804 & \underline{43.6} & 88.7\% & \underline{0.895} & \underline{36.0} & 97.1\% & \underline{0.890} & \underline{44.9} & 93.0\% & 0.795 & \underline{54.1} & 80.7\% \\
    GeoTikzBridge-Base-38B & \bf 0.813 & \bf 39.7 & \bf 95.1\% & \bf 0.915 & \bf 30.6 & 95.2\% & \bf 0.900 & \bf 40.7 & \bf 95.8\% & \bf 0.821 & \bf 47.6 & \bf 83.4\% \\
    \bottomrule
  \end{tabular}
\end{table*}

\begin{table*}[t]
\renewcommand{\arraystretch}{0.8}
\footnotesize
\vspace{-5pt}
  \caption{Performance comparison of downstream mathematical reasoning across five benchmarks. The best and second-best results are marked with bold and underlines, respectively.}\vspace{-5pt}
  \label{table: LLMs_VLMs_TikZ}
  \centering
  \begin{tabular}{p{6.5cm}>{\centering\arraybackslash}p{1.5cm}>{\centering\arraybackslash}p{2.0cm}>{\centering\arraybackslash}p{1.5cm}>{\centering\arraybackslash}p{1.5cm}>{\centering\arraybackslash}p{1.5cm}}
    \toprule
    \multirow{2}{*}{Method} & MathVista & GAOKAO-MM & RBench-M & RBench-V & RBench-V \\
    & (GPS) & (Math) & (en) & (Math) & (Overall) \\
    \midrule
    \rowcolor{gray!30}
    \multicolumn{6}{c}{\textit{Closed-sourced VLMs}} \\
    \midrule
    GPT-4V\cite{openai2023gpt4v} & 0.505 & 0.450 & - & - & - \\
    GPT-4-turbo\cite{achiam2023gpt} & 0.583 & 0.500 & - & - & - \\
    GPT-4o\cite{hurst2024gpt} & 0.647 & - & 0.334 & 0.244 & 0.141 \\
    Claude3.5-Sonnet\cite{anthropic2024claude35sonnet} & 0.644 & - & 0.397 & - & - \\
    Gemini-1.5-Pro\cite{team2024gemini} & 0.589 & - & 0.355 & - & - \\
    \midrule
    \rowcolor{gray!30}
    \multicolumn{6}{c}{\textit{Open-sourced VLMs (Base or Base/\bf{ours})}} \\
    \midrule
    Qwen2.5-VL-32B\cite{Qwen2.5-VL} & 0.654 & 0.500 & 0.359 & 0.227 & 0.100 \\
    Qwen2.5-VL-72B\cite{Qwen2.5-VL} & 0.644 & 0.613 & 0.304 & 0.153 & 0.106 \\
    InternVL3-78B\cite{zhu2025internvl3} & 0.663 & 0.638 & 0.397 & 0.159 & 0.101 \\
    Qwen3-VL-30B-A3B\cite{qwen3technicalreport}/+GeoTikzBridge-Base & 0.697/0.745 & 0.550/0.588 & 0.314/0.320 & 0.218/0.236 & 0.110/0.117 \\
    InternVL3.5-38B\cite{wang2025internvl3_5}/+GeoTikzBridge-Base & 0.688/0.718 & 0.600/0.635 & 0.353/0.369 & 0.216/0.232 & 0.118/0.120 \\
    GLM4.5-V-106B-A12B\cite{vteam2025glm45vglm41vthinkingversatilemultimodal}/+GeoTikzBridge-Base & 0.745/0.764 & 0.613/0.663 & 0.434/\bf{0.498} & 0.230/\underline{0.295} & 0.125/0.133 \\
    \midrule
    \rowcolor{gray!30}
    \multicolumn{6}{c}{\textit{LLMs + Tikz ({\bf ours})}} \\
    \midrule
    Skywork-OR1-32B\cite{he2025skywork} + GeoTikzBridge-Base & 0.861 & 0.663 & \underline{0.436} & 0.253 & \underline{0.150} \\
    Qwen3-30B-A3B\cite{yang2025qwen3} + GeoTikzBridge-Base & 0.856 & 0.550 & 0.397 & 0.267 & 0.128 \\
    GLM-Z1-32B\cite{glm2024chatglm} + GeoTikzBridge-Base & 0.875 & \underline{0.675} & 0.305 & 0.201 & 0.105 \\
    DS-R1-Distill-Qwen-32B\cite{deepseekai2025deepseekr1incentivizingreasoningcapability} + GeoTikzBridge-Base & 0.841 & \bf{0.688} & 0.364 & 0.213 & 0.112 \\
    GLM-4.5-Air-106B-A12B\cite{zeng2025glm} + GeoTikzBridge-Base & \bf 0.889 & \bf{0.688} & 0.359 & 0.250 & 0.138 \\
    GPT-OSS-120B\cite{agarwal2025gpt} + GeoTikzBridge-Base & \underline{0.880} & 0.625 & 0.430 & \bf 0.301 & \bf 0.164 \\
    \bottomrule
  \end{tabular}
  \vspace{-5pt}
\end{table*}

\begin{table*}[t]
\renewcommand{\arraystretch}{0.8}
\footnotesize
  \caption{Performance comparison of instructed code generation in the downstream auxiliary line generation task. The best results are marked with bold.}\vspace{-5pt}
  \label{table: auxiliary_line_addition}
  \centering
  \begin{tabular}{p{3.0cm}>{\centering\arraybackslash}p{1.8cm}>{\centering\arraybackslash}p{1.8cm}>{\centering\arraybackslash}p{1.8cm}>{\centering\arraybackslash}p{1.8cm}>{\centering\arraybackslash}p{1.8cm}>{\centering\arraybackslash}p{1.8cm}}
    \toprule
    Method & MSE$\downarrow$ & SSIM$\uparrow$ & PSNR$\uparrow$ & CLIP Score$\uparrow$ & FID$\downarrow$ & CSR$\uparrow$ \\
    \midrule
    FigCodifier\cite{wang2025mathcoder} & 1435.9 & 0.135 & 17.6dB & 0.936 & 9.798 & 88.0\% \\
    GeoTikzBridge-Base-8B & 692.6 & 0.549 & 23.1dB & 0.967 & 5.476 & 95.4\% \\
    GeoTikzBridge-Instruct & \bf 211.7 & \bf 0.844 & \bf 89.6dB & \bf 0.992 & \bf 1.158 & \bf 96.7\% \\
    \bottomrule
  \end{tabular}
\end{table*}


\begin{table}[t] 
\renewcommand{\arraystretch}{0.8} 
\footnotesize 
\vspace{-5pt}
\caption{Performance comparison of four different visual reasoning formats on the MathVista-GPS benchmark. The baseline model is InternVL3.5-38B.}
\vspace{-5pt}
\label{table: AuxLine_PS}
\centering
\begin{tabular}{
  p{3.9cm}
  >{\centering\arraybackslash}p{0.6cm} 
  >{\centering\arraybackslash}p{0.6cm} 
  >{\centering\arraybackslash}p{0.6cm} 
  >{\centering\arraybackslash}p{0.6cm} 
}
\toprule

Exp ID & I & II & III & IV \\ 
\midrule
Problem Text & $\checkmark$ & $\checkmark$ & $\checkmark$ & $\checkmark$ \\
Raw Geometric Image & $\checkmark$ & $\checkmark$ & $\checkmark$ & $\checkmark$ \\
Tikz Code With Aux. Line Added & $\times$ & $\times$ & $\checkmark$ & $\checkmark$ \\
Image With Aux. Line Added & $\times$ & $\checkmark$ & $\times$ & $\checkmark$ \\
Answer Accuracy & 0.688 & 0.697 & 0.707 & \bf 0.736 \\
\bottomrule
\end{tabular}
\end{table}

\section{Experiments}
\label{sec:exp}
In this section, we introduce the experimental results in both image-to-tikz and downstream visual reasoning tasks.

\subsection{Settings}


\textbf{Implementation Details.} We develop two GeoTikzBridge-Base models, initialized from pretrained FigCodifier-8B and InternVL3.5-38B-Instruct respectively. For the 8B model, full-parameter SFT is adopted with learning rate of 4e-7, batch size per device of 4, and gradient accumulation steps of 4. For the 38B model, LoRA fine-tuning is applied to all QKV layers, using learning rate of 1e-4, batch size per device of 1, and gradient accumulation steps of 4. DeepSpeed (ZeRO-3 stage\cite{rajbhandari2020zero}) and flash attention\cite{dao2022flashattention} are employed. Both models are trained for 1 epoch on 8 NVIDIA H100 GPUs (80GB). For reproducibility, greedy decoding with temperature 0 is used to generate tikz codes during testing, and the threshold $\tau$ for selecting reliable generated code is set to 0.8. 
To advance geometric visual reasoning, we further introduce a model that edits tikz code to add auxiliary lines, which is fine-tuned from GeoTikzBridge-Base-8B for 3 epochs with a learning rate of 4e-7.
Training costs are $\sim$96 GPU hours (8B) and $\sim$488 GPU hours (38B). The choice of $K=4$ is analyzed in Sec. E of supplementary.

\textbf{Benchmarks and Metrics.} We evaluate our models on five open-sourced benchmarks (DaTikZ\cite{belouadi2025tikzero}, MathVista-GPS\cite{lu2023mathvista}, GAOKAO-MM-Math\cite{zong2024gaokao}, R-Bench\cite{rBench}, RBench-V\cite{rBench-V}) and two in-house benchmarks (EDUBenchmark and GeoTikz-Instruct), focusing on mathematical visual reasoning. Following\cite{rodriguez2025starvector, yang2025omnisvg}, we adopt CLIP Score\cite{radford2021clip} and FID\cite{radford2021clip} for image-to-tikz task, along with fine-grained scores including MSE, SSIM, and PSNR for image-to-auxiliary-line task. Besides, we introduce a stricter metric, Compilation Success Rate, which is defined as the percentage of codes that can be successfully compiled and produce valid non-blank images, since most existing methods neglect the compilation errors that may occur in the code-to-image process. Details are introduced in Supplementary Material.

\subsection{Evaluation}

\textbf{Evaluation on Image-to-Tikz.} Table \ref{table: GeoTikzBridge-Base_performance} presents CLIP scores (CLIP-S), FID, and Compilation Success Rates (CSR) of models across four benchmarks in the GeoTikzBridge-Base task. As shown in Table \ref{table: GeoTikzBridge-Base_performance}, our proposed GeoTikzBridge-Base models achieve state-of-the-art (SOTA) performance in the GeoTikzBridge-Base task in most cases.
Notably, GeoTikzBridge-Base-38B achieves the best performance across all benchmarks, as measured by the CLIP Score, significantly outperforming 6 open-sourced VLMs and the existing image-to-Tikz model.

\textbf{Evaluation on LLMs/VLMs Multimodal Reasoning.} We evaluate the performance of various LLMs and VLMs on multimodal reasoning tasks when provided with tikz codes generated by our GeoTikzBridge-Base-8B model. The results are summarized in Table \ref{table: LLMs_VLMs_TikZ}, from which three main points can be observed: (i) Models provided with tikz codes from our GeoTikzBridge-Base achieve the best performance in most cases, demonstrating the effectiveness of our tikz-enhanced reasoning method; (ii) Supplementing with tikz codes significantly boosts VLMs' performance, indicating that symbolic geometric representations can compensate for VLMs' limitations in visual perception and understanding; (iii) LLMs provided with tikz codes generally outperform their VLM counterparts under the same setting. This can be attributed to catastrophic forgetting: When VLMs are trained on visual-language alignment, their inherent language capabilities often degrade. 


\textbf{Evaluation on Instructed Code Generation.} We evaluate our GeoTikzBridge-Instruct in generating tikz code from geometric figures with specified instructions, e.g., ``Draw a line through point C parallel to AB". The model takes raw geometric images and auxiliary line instructions as input, outputting tikz codes for figures with added auxiliary lines. 
As shown in Table \ref{table: auxiliary_line_addition}, GeoTikzBridge-Instruct achieves strong performance across multiple metrics, demonstrating that it can capture the geometric semantic and structural details more accurately.


\textbf{Evaluation on Problem Solving With Auxiliary Line.} We evaluate how auxiliary lines from our GeoTikzBridge-Instruct affect geometric problem solving with different visual reasoning formats. Firstly, InternVL3.5-38B predicts auxiliary line instructions for original geometric problems, excluding those unnecessary for auxiliary lines. Then, these instructions and raw images are fed into GeoTikzBridge-Instruct to generate tikz codes with added auxiliary lines. Finally, the generated tikz codes and/or rendered images are fed into InternVL3.5-38B for mathematical reasoning. The corresponding results on the MathVista-GPS benchmark are reported in Table \ref{table: AuxLine_PS}. Two points can be seen from this table: (i) Adding auxiliary lines is helpful for problem solving, no matter in the form of rendered image or tikz code; (ii) Tikz code is still more effective than image in this task.

\begin{figure}[t]
  \centering
    \includegraphics[width=6.8cm, height=4.6cm]{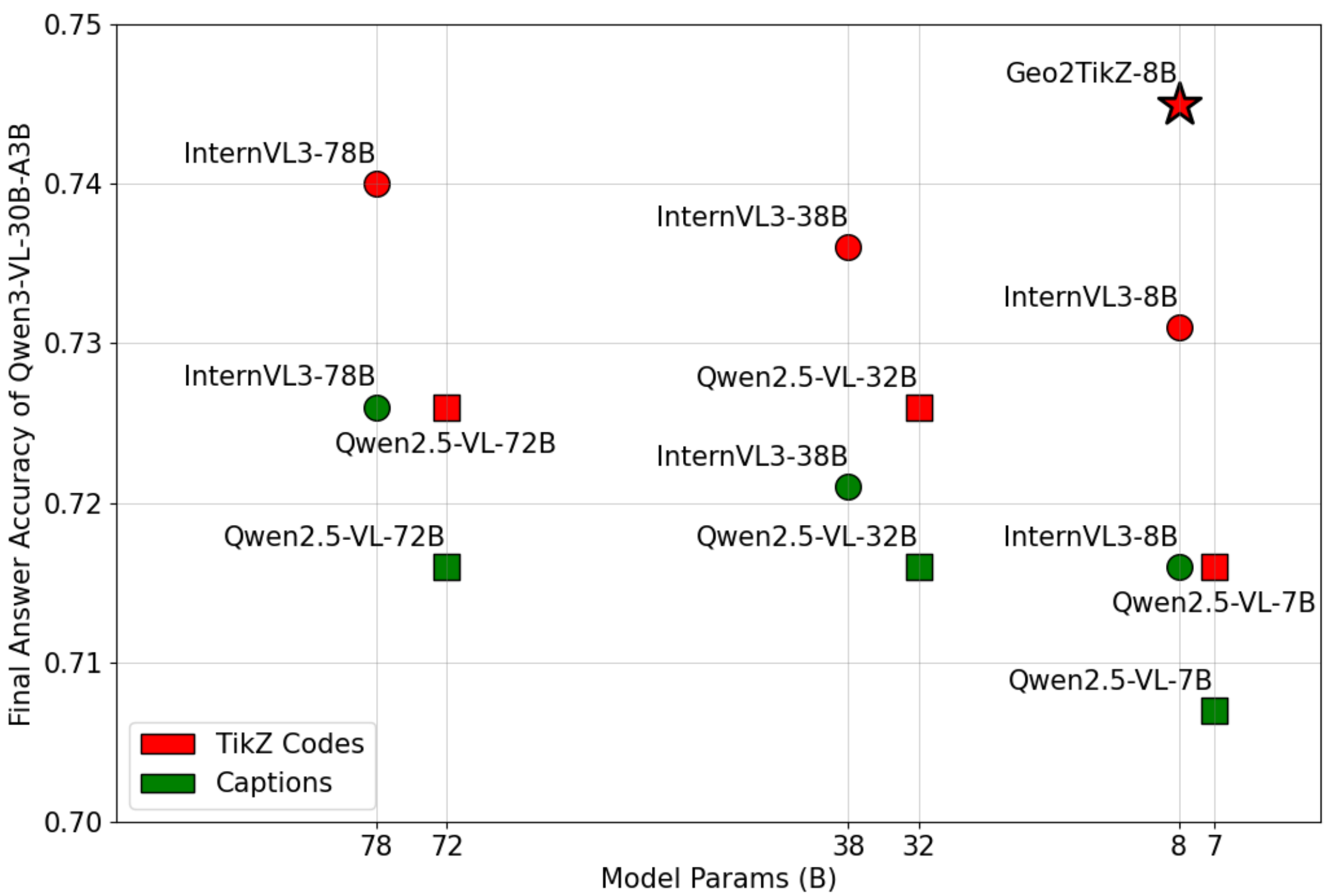}
    \vspace{-5pt}
    \caption{Ablation study of geometric representation on the MathVista-GPS benchmark. The baseline model is Qwen3-VL-30B-A3B.}\vspace{-12pt}
    \label{fig: ablation_different_texts}
\end{figure}

\begin{figure}[t]
  \centering
    \includegraphics[width=8.0cm, height=4.5cm]{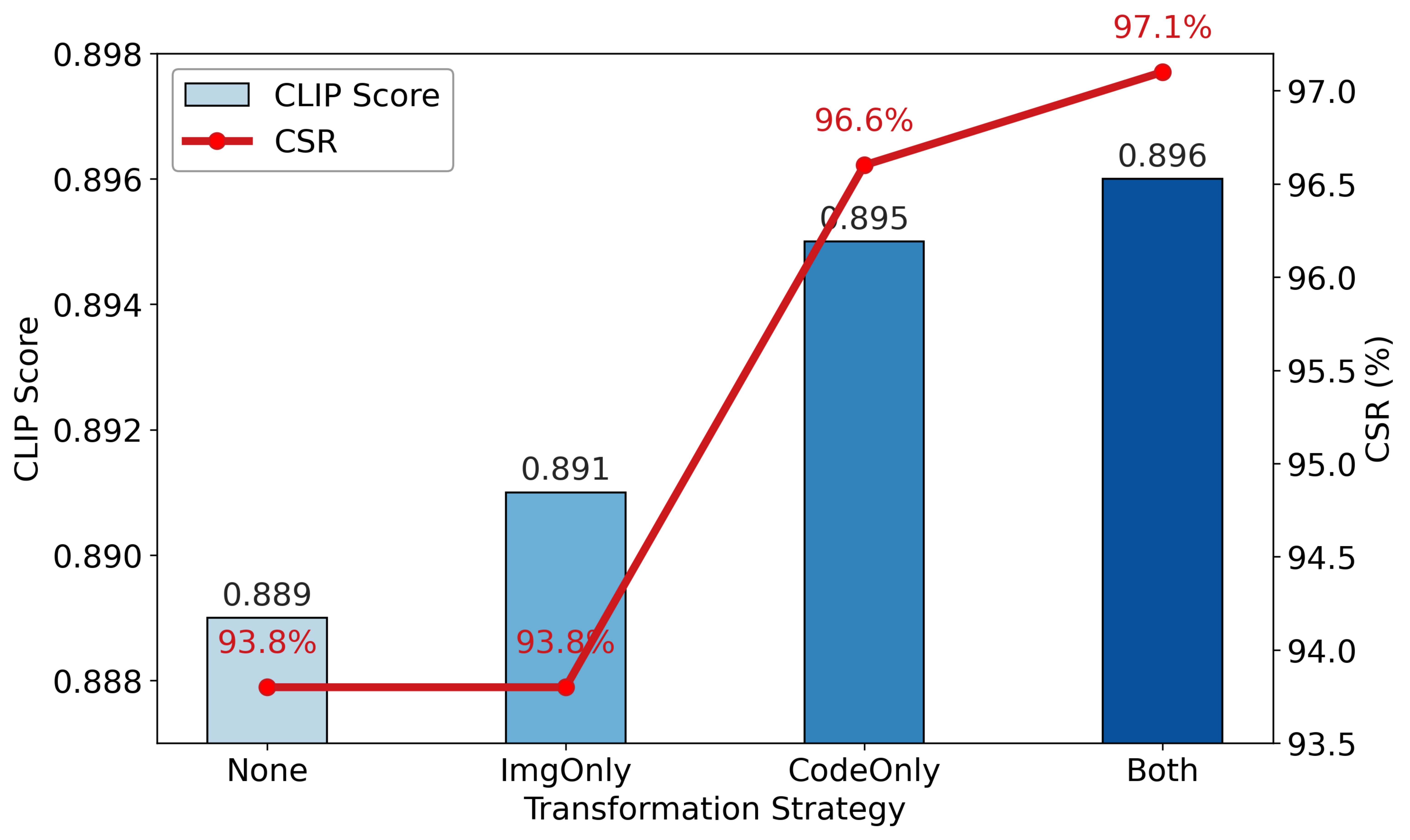}
    \vspace{-5pt}
    \caption{Ablation study of localized geometric transformation strategies for GeoTikzBridge-Base-8B, evaluated on the MathVista-GPS benchmark.}\vspace{-10pt}
    \label{fig: ablation_augmentations}
\end{figure}

\subsection{Ablations}

\begin{figure*}[t]
  \centering
    \includegraphics[width=16.2cm, height=5.0cm]{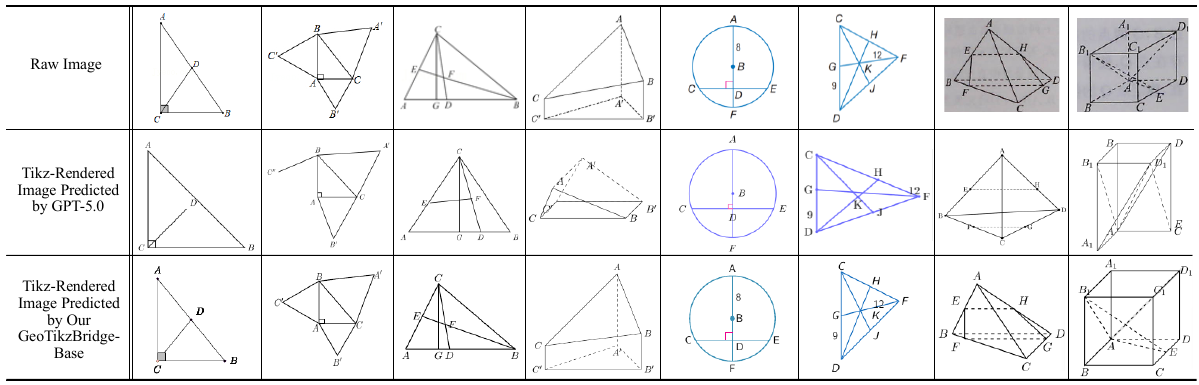}
    \vspace{-8pt}
    \caption{Visualization results of tikz codes generated by GeoTikzBridge-Base compared with GPT-5.0\cite{GPT5}, each figure shows the input image (either geometric or non-geometric) and the corresponding image rendered from our predicted tikz code.}
    \vspace{-8pt}
    \label{fig: GeoTikzBridge-Base_results}
\end{figure*}

\begin{figure*}[t]
  \centering
    \includegraphics[width=16.2cm, height=5.5cm]{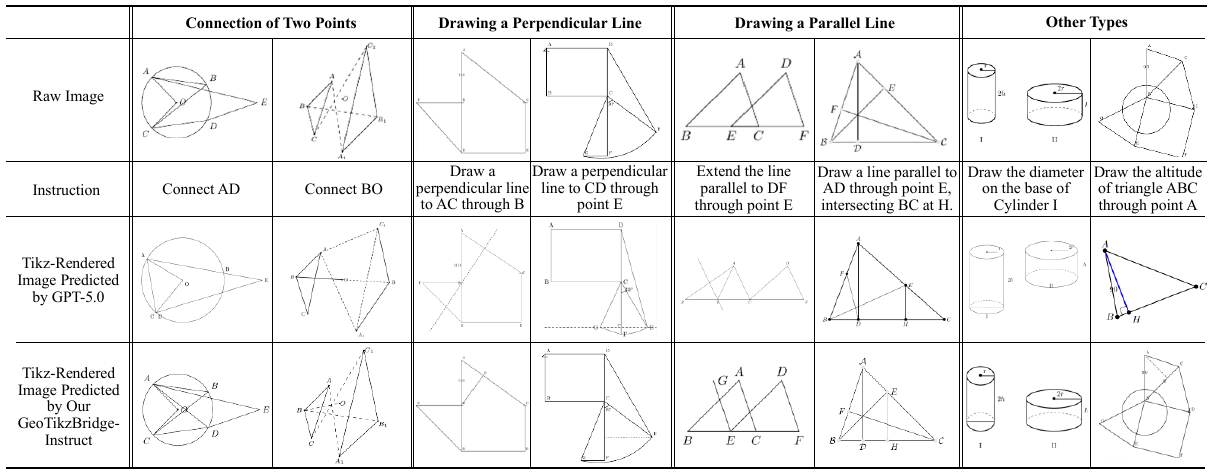}
    \vspace{-8pt}
    \caption{Visualization results of instruction-driven auxiliary line generation compared with GPT-5.0\cite{GPT5}, each figure shows the input geometric image, the natural language instruction for auxiliary line addition, and the rendered image from the generated tikz code.}
    \vspace{-8pt}
    \label{fig: aux_line_results}
\end{figure*}

\textbf{Geometric Representations in Reasoning Task.} 
To investigate how different geometric representations influence the geometric reasoning performance of models, we selected two typical types of representations for comparative analysis: one is natural language captions generated by open-sourced VLMs (Qwen2.5-VL and InternVL3 series), and the other is tikz codes produced by the same VLMs and the proposed GeoTikzBridge-Base-8B model. All these representation forms need to convert original geometric images into geometric representations, which are then input into a specific VLM (Qwen3-VL-30B-A3B) together with problem texts and original images for reasoning.

Fig. \ref{fig: ablation_different_texts} presents the ablation results on the MathVista-GPS benchmark. The results indicate that, for the same model, the generated tikz codes can drive VLM to handle geometric reasoning more effectively than natural language descriptions. This fully demonstrates the advantages of tikz codes in representing geometric information. Meanwhile, the tikz codes generated by the proposed GeoTikzBridge-Base-8B model contribute more significantly to the improvement of VLMs' geometric reasoning performance compared with the tikz codes generated by models with larger parameter sizes, demonstrating the effectiveness of our method.


\textbf{Effect of Localized Geometric Transformation.} We analyze the effectiveness of the proposed localized geometric transformation strategy, including image and code transformation. Experiments are conducted using GeoTikzBridge-Base-8B on the MathVista-GPS benchmark, with results presented in Fig. \ref{fig: ablation_augmentations}. As shown in this figure, the code transformation strategy achieves substantial gains in both CLIP score and compilation success rate, primarily due to its capacity to not only reduce redundant code generation but also enhance geometric details via localized code editing.

\subsection{Visualization}

We present generated examples for the image-to-tikz and image-to-auxiliary-line tasks, with comparative analysis against GPT-5.0\cite{GPT5} conducted for each task. As illustrated in Fig. \ref{fig: GeoTikzBridge-Base_results}, our model more precisely captures key geometric components, including line segments, angles, closed polygons, and semantic labels. Fig. \ref{fig: aux_line_results} demonstrates that our model outperforms GPT-5.0 across diverse geometric problem scenarios: It generates auxiliary lines more compliant with given instructions while maintaining better preservation of the original geometric figure characteristics. Additional results are provided in the Supplementary Material.

\section{Conclusion}
\label{sec:conclusion}
In this paper, we present GeoTikzBridge, a framework that enhances geometric perception reasoning in MLLMs and LLMs. Within this framework, we introduce two datasets. GeoTikz-Base is the largest image-to-tikz dataset to date, containing 2.5M pairs—16× larger than existing open-source datasets. We also present GeoTikz-Instruct, the first instruction-augmented tikz dataset designed for visual reasoning.
Trained on these datasets, GeoTikzBridge models achieve state-of-the-art performance among open-source MLLMs. They can also serve as plug-and-play modules to enhance geometric reasoning in VLMs and LLMs without extensive retraining. The fine-tuned model further boosts MLLMs' reasoning capabilities through instructed code generation, bridging the gap between visual understanding and symbolic reasoning.


In the future, we will expand this approach to cover technical diagrams and schematics, including engineering circuit diagrams and advanced educational reasoning problems. Additionally, we aim to explore integrating GeoTikzBridge as a code-generation tool for multimodal agents in an Agentic RL manner to further enhance reasoning capabilities.


{
	\small
	\bibliographystyle{ieeenat_fullname}
	\bibliography{main}
}
\clearpage
\setcounter{page}{1}
\renewcommand\thesection{\Alph{section}}
\setcounter{figure}{0}
\renewcommand*{\thefigure}{S\arabic{figure}}
\renewcommand*{\thetable}{S\arabic{table}}
\maketitlesupplementary

This supplementary material complements the main text: Firstly, Sec. \ref{sec: supple-A} specifies seven benchmarks (five open-sourced, two in-house constructed) for geometric perception and reasoning, along with seven metrics: CLIP Score, FID, and Compilation Success Rate (CSR) for image-to-tikz generation; MSE, SSIM, PSNR for fine-grained auxiliary line addition; and Final Answer Accuracy (top-1 accuracy) for downstream reasoning. Secondly, Sec. \ref{sec: supple-B} covers compilation details, including a post-processing workflow for syntax and logical errors and LaTeX compilation-rendering process. Thirdly, Sec. \ref{sec: supple-C} provides Doubao prompt for filtering unreliable instruction annotations in constructing the GeoTikz-Instruct dataset. Then, Sec. \ref{sec: supple-D} describes the image transformation strategy to enrich training image diversity. Next, Sec. \ref{sec: supple-E} presents ablation studies analyzing impacts of maximum self-refinement iteration number $K$ and the threshold $\tau$ in reliable code selection. Then, we provide failure analysis in Sec. \ref{sec: supple-F} and analyze the model-agnostic effectiveness of our approach in Sec. \ref{sec: supple-G}, respecitvely. Next, we supple quantitive comparison results with latest closed-source models in Sec. \ref{sec: supple-H}. Finally, Sec. \ref{sec: supple-I} provides supplementary visualization results, including additional examples of image-to-tikz generation, instruction-driven auxiliary line generation, as well as tikz assistant and auxiliary line assistant reasoning.

\section{Benchmark \& Metric Details}
\label{sec: supple-A}
\subsection{Benchmark Details}
five open-sourced benchmarks (i.e., DaTikZ, MathVista-GPS, GAOKAO-MM-Math, RBench, and RBench-V) and two in-house constructed benchmarks named EDUBenchmark and the constructed GeoTikz-Instruct are used for evaluation, most of which are focused on geometric mathematical reasoning.

\begin{itemize}
  \item \textit{DaTikZ} is a large-scale image-tikz benchmark (360k+ graphics) for scientific diagram analysis, which requires LaTeX code parsing and text-graphics mapping. The GeoTikzBridge-Base results are reported on its test set, which contains 542 graphic images.
  \item \textit{MathVista-GPS} focuses on visual-geometric reasoning, which is a subset of MathVista benchmark. It contains 208 samples covering 2D/3D geometry and dynamic transformations. It includes diverse visual contexts, such as natural images, geometric diagrams, abstract scenarios, synthetic scenes, and various charts and figures.
  \item \textit{GAOKAO-MM-Math} is a subset of GAOKAO-MM with 80 questions that focus on mathematical reasoning and is a Gaokao-based benchmark. The GeoTikzBridge-Base results are reported on its 142 images.
  \item \textit{RBench} is a graduate-level multi-disciplinary benchmark designed to evaluate the complex reasoning capabilities of MLLMs/LLMs, spanning 19 departments (e.g., mathematics, physics, biology) and over 100 subjects. It supports both English and Chinese, with 665 multimodal-specific questions integrating text and visual modals. We adopt its ``Multimodal English (en)" subset (665 samples) for experiments.
  \item \textit{RBench-V} is a vision-indispensable reasoning benchmark covering 4 core categories: math, physics, counting, and game. It contains 803 multimodal questions. In our work, we focus on two categories: the ``Overall" category (totally 803 samples) for general reasoning, and the ``Math" category (176 samples) for geometric reasoning.
  \item \textit{EDUBenchmark} is an in-house constructed benchmark for multi-disciplinary reasoning. It contains 373 graphic images covering multiple disciplines (such as math, physics, chemistry, etc.) across age groups corresponding to primary school, junior high school, senior high school, university, and vocational exams, ensuring comprehensive coverage of multimodal mathematical reasoning scenarios.
  \item \textit{GeoTikz-Instruct} is an in-house constructed benchmark for adding auxiliary lines in geometric figures, as mentioned before. It consists of 419k training samples and 789 testing samples.
\end{itemize}

\subsection{Evaluation Metric Details}
The image-to-tikz task focuses on generating valid tikz codes from geometric figures, so we adopt two metrics to measure both graphical consistency and code functionality:
\begin{itemize}
\item \textit{CLIP Score}: We use the CLIP model with ViT-L/14@224px as vision encoder to quantify semantic consistency between the image that rendered from the generated tikz code and the original image. CLIP score is calculated as the cosine similarity between the two image embeddings.
\item \textit{FID}: Fréchet Inception Distance (FID) is also utilized for generated tikz quality evaluation. ViT-L/14@224px is re-employed to extract features from rendered and original images. FID computes the mean and covariance of the two feature distributions, and calculates the Fréchet distance between the corresponding multivariate Gaussians. Smaller values indicate higher distribution similarity and better generation quality.
\item \textit{Compilation Success Rate}: We evaluate the functional correctness of the generated tikz code by compiling and rendering. The metric is defined as the percentage of code samples that successfully compile without errors and produce non-blank images.
\end{itemize}

Compared with geometric tikz code generation, the auxiliary line addition task is a more fine-grained task, as it requires precise localization of geometric primitives (e.g., vertices, line segments) and the generation of logically consistent auxiliary structures to connect known conditions, rather than merely reproducing the overall graphical layout. Hence, besides CLIP score and compilation success rate, we also adopt three fine-grained metrics to assess the similarity between each rendered image from the generated tikz and the corresponding ground-truth image in the test set:
\begin{itemize}
\item \textit{MSE}: Mean Square Error (MSE) is a fundamental metric for quantifying pixel-wise differences between images rendered from generated tikz codes and ground-truth images. It is calculated as the average of squared errors between corresponding pixels of the two images. Lower MSE values indicate smaller pixel-level discrepancies and higher rendering fidelity.
\item \textit{SSIM}: Structural Similarity Index (SSIM) aligns with human visual perception by incorporating three key attributes of images: luminance, contrast, and local structure. It computes similarity between images rendered from generated tikz codes and ground-truth images through sliding window-based regional pixel analysis, with value ranging from 0 to 1. Values closer to 1 indicate greater structural and visual consistency between the two images.
\item \textit{PSNR}: Peak Signal to Noise Ratio (PSNR) is derived from MSE and measured in decibels (dB) to evaluate image quality by quantifying the ratio between signal peak amplitude and noise. It compares images rendered from generated tikz codes against ground-truth images, with higher PSNR values corresponding to less distortion.
\end{itemize}


In downstream mathematical reasoning task, \textit{Final Answer Accuracy} is adopted, which is defined as the correctness of the generated final solution.

\section{Compilation Details}
\label{sec: supple-B}


\textbf{Code Post-processing.} Given that tikz codes predicted by the model are prone to format non-compliance, leading to compilation failures, a code post-processing strategy is implemented to fix common compilation errors. This integrated workflow combines syntax correction and logical optimization, mainly addressing unclosed environments and redundancy issues, which not only enhances code compileability, but also improves the compactness and readability of the codes. Notably, the reported experimental results have employed the same post-processing operations for all models, including our proposed models and the compared models. The post-processing procedure is as follows:
\begin{enumerate}[1)]
\item \textit{Package Deduplication:} For duplicate `\textbackslash usepackage' declarations, retain one instance per unique package while preserving all non-redundant statements.
\item \textit{Block-Aware Deduplication:} Parse and identify nested environments (e.g., `tikzpicture', `scope') with priority to innermost environments as well as `\textbackslash foreach' code blocks. Within each identified block, remove redundant identical non-environment start/end lines, avoiding redundant code execution while maintaining structural integrity.  
\item \textit{Truncated Line Removal:} Detect syntactically incomplete lines, defined as lines with unclosed LaTeX commands (excluding environment directives) lacking terminating semicolons, or imbalanced curly braces/square brackets, and discard only the last non-symbol-only line (if incomplete) before the main environment end `\textbackslash end\{tikzpicture\}'.
\item \textit{Environment Balance:} Adopt stack-based nesting tracking to manage nested environments. Append missing environment end directives in reverse order of their opening (e.g., close `scope' before `tikzpicture' for nested structures).  
\item \textit{Document Structure Optimization:} Ensure the `document' environment is fully defined with exactly one pair of start (`\textbackslash begin\{document\}') and end (`\textbackslash end\{document\}') directives, appending `\textbackslash end\{document\}' at the end (after removing trailing blank lines) if absent. Relocate any graphical environments (e.g., `tikzpicture') erroneously placed outside the `document' scope into it for syntactic validity.  
\item \textit{Logical Refinement:} Extract redundant graphical commands (e.g., `\textbackslash draw', `\textbackslash node') from `\textbackslash foreach' blocks that do not reference the loop variable (e.g., `\textbackslash x', `\textbackslash y'), relocating them outside the loop to avoid redundant execution. Besides, move commands misplaced outside their semantically required environments (e.g., `\textbackslash draw' outside `tikzpicture') to the correct scope.  
\end{enumerate}
After post-processing, the Compilation Success Rate (CSR) metrics of generated codes reflect the intrinsic code generation quality and syntax adaptation potential of different methods, highlighting inherent model capability in terms of code syntax correctness, structural rationality, and format compliance, eliminating interference from non-essential factors such as sequence disorder and redundant repetition. 

\textbf{Compiling and Rendering.} Each post-processed tikz code snippet is embedded within a LaTeX code framework to enable standardized rendering of its defined geometric content. We also include the necessary language packages at the beginning of the LaTeX file to ensure that the text content in the corresponding languages can be displayed successfully in the compiled PDF. Besides, we set the LaTeX file type to `standalone' to facilitate subsequent compiling and rendering into images. Then, the latex codes are compiled into PDFs, which are subsequently converted to Portable Network Graphics (PNG) images via the pdf2Image library.


\section{Doubao Prompt for Filtering Unreliable Instruction Annotations} 
\label{sec: supple-C}

\begin{figure}[h]
  \centering
    \includegraphics[width=8.2cm, height=8.5cm]{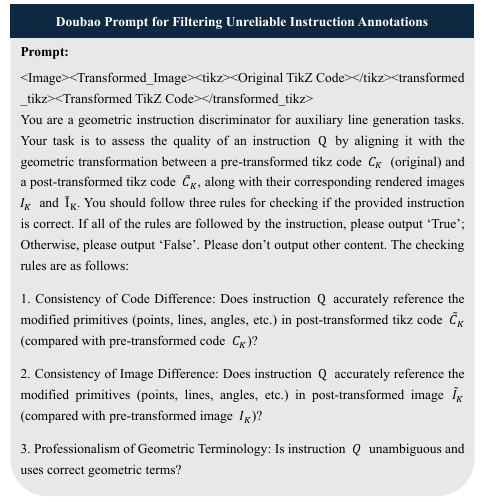}
    \caption{Prompt input to Doubao for filtering unreliable instruction annotations in the GeoTikz-Instruct dataset construction.}
    \label{fig: prompt_for_Doubao}
\end{figure}

To filter out low-quality GeoTikz-Instruct instructions annotated by Qwen2.5-VL-72B while ensuring geometric validity, we adopt a prompt-guided discriminative strategy using Doubao. Specifically, a targeted prompt is first designed for guiding Doubao to score the raw annotated instructions based on geometric consistency, as shown in Fig. \ref{fig: prompt_for_Doubao}. Then, high-quality outputs are selected to form the final instruction-image-tikz triples.

\section{Image Transformation Details}
\label{sec: supple-D}

As a key component of the localized geometric transformation strategy (introduced in Sec. 3.2 of the main text), image transformation complements code transformation to address the model's insensitivity to fine-grained geometric details. While code transformation enhances structural semantics via localized editing, fine-grained visual variations (e.g., slight rotations, scale adjustments) in geometric scenes still need handling. The core goal of image transformation is to enrich the visual diversity of training samples in images while preserving invariant geometric relationships in codes, forcing the model to focus on essential geometric properties rather than superficial visual features. The image transformation procedure is as follows:

\begin{enumerate}[1)]
\item \textit{Canvas Expansion}: A 100-pixel replication border is appended around the original image to form an expanded canvas, which reserves sufficient edge space to prevent content truncation induced by subsequent perspective transformation.
\item \textit{Perspective Transformation}: Source points are defined as the four corners of the original image on the expanded canvas, with target points generated through random offsets (ranging from -50 to 50 pixels, equivalent to half the border size). A perspective transformation matrix is estimated based on source-target correspondences and applied to the canvas to introduce subtle subtle distortion that simulates natural viewing angle variations.
\item \textit{Optimized Cropping}: The transformed image is converted to gray-scale, binarized, and processed using morphological closing with a 15$
\times$15 kernel to fill holes in the foreground region. The maximum external contour is selected to compute the minimum enclosing rectangle, and the cropping boundary is expanded by 5 pixels. To preserve key content, the cropped region is constrained to be at least 50\% of the original image size. Subsequently, the cropped image is resized to the original dimensions using area interpolation to ensure sharp down-sampling.
\item \textit{Illumination Enhancement}: Random non-uniform illumination gradients are introduced, including horizontal (with brightness linearly transitioning from 0.4 to 0.6 on the left to 0.8 to 1.0 on the right), vertical (with brightness linearly transitioning from 0.4 to 0.6 at the top to 0.8 to 1.0 at the bottom), and central (with brightness decaying from the center to the edges). The central gradient follows$\text{gradient}(x,y) = \text{base\_brightness} + \text{brightness\_range} \times (1 - \frac{\text{dist}}{\text{max\_dist}})$, where $\text{base\_brightness}$ ranges from 0.6 to 0.7, $\text{brightness\_range}$ ranges from 0.2 to 0.3, dist denotes the distance from a pixel to the center, and $\text{max\_dist}$ is the maximum diagonal distance from the center. The gradient matrix is multiplied with the image, and the resulting pixel values are clipped to the range [0, 255].
\item \textit{Blur Augmentation}: Gaussian blur is applied with a 50\% probability, using a radius ranging from 0.5 to 1.2 to simulate mild lens defocus.
\item \textit{Color Adjustment}: Contrast (with a factor ranging from 0.7 to 0.8) and saturation (with a factor ranging from 0.7 to 0.85) are moderately reduced to mimic natural color attenuation.
\item \textit{Lens Distortion}: Subtle radial distortion is introduced, with distortion coefficients $k_1$ in the range [-0.03, 0.03] and $k_2$ in [-0.01, 0.01]. Image coordinates are normalized to the range [-1, 1], and the radial distance is computed as $r = x_{\text{norm}}^2 + y_{\text{norm}}^2$. Distorted coordinates are derived as $x' = x_{\text{norm}} \times (1 + k_1 r + k_2 r^2)$ and $y' = y_{\text{norm}} \times (1 + k_1 r + k_2 r^2)$, which are then remapped to the original image size.
\item \textit{Rotation Augmentation}: Minor in-plane rotation (within the range $[-1.5^\circ, 1.5^\circ]$) is applied around the image center, with a replication border mode used to handle edge pixels.
\end{enumerate}

\section{Supplementary Ablations}
\label{sec: supple-E}

\begin{figure}[t]
  \centering
    \includegraphics[width=7.6cm, height=4.6cm]{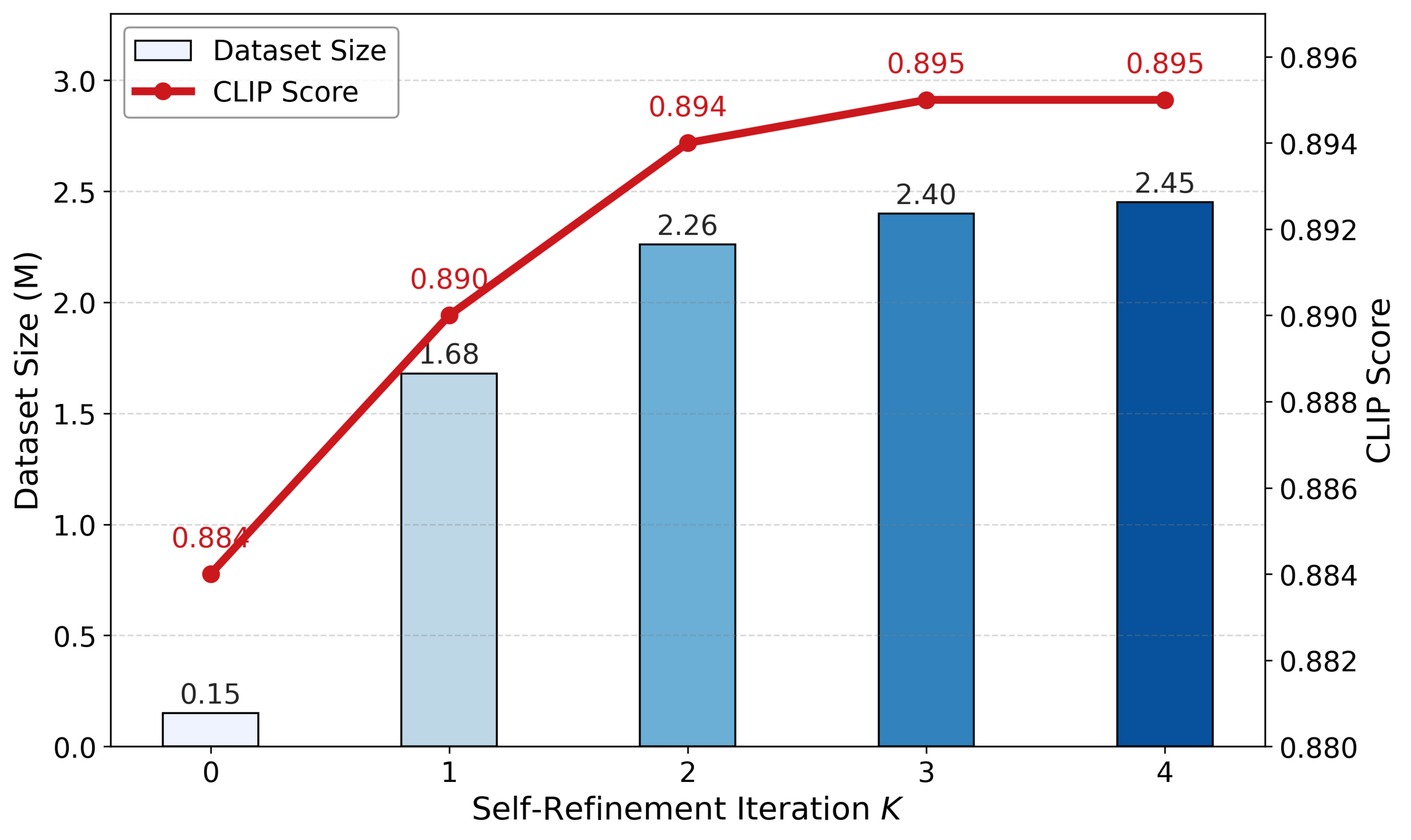}
    \caption{Ablation study of different values of self-refinement $K$ for GeoTikzBridge-Base-8B, evaluated on the MathVista-GPS benchmark.}
    \label{fig: ablation_on_K}
\end{figure}

\textbf{Self-Refinement Iteration $K$.} We evaluate the effect of iterative self-refinement on the image-to-tikz performance of GeoTikzBridge-Base-8B and the scalability of its training dataset, Fig. \ref{fig: ablation_on_K} presents the corresponding ablation study. Evaluated on the MathVista-GPS benchmark, the figure uses bars to represent training dataset size and lines to denote CLIP score.  

$K=0$ corresponds to the initial state where the pre-trained FigCodifier (denoted as $M_0$) utilizes the 145k-sample DaTikZ seed dataset. This model achieves a CLIP score of 0.884 on geometric perception, as it has not been trained on geometric-specific dataset. At $K=1$, $M_0$ is employed to predict tikz codes for candidate geometric images sourced from nine public datasets. The limited ability of $M_0$ to parse fine-grained details (e.g., segment relations, angles) results in relatively few reliable samples that are filtered via the CLIP threshold of 0.8 (detailed in Sec. 3.2 of the main text). With this expanded dataset, the model trained on it (denoted as $M_1$) outperforms $M_0$ in CLIP score, as the added geometric-specific training samples enhance basic geometric perception of the model.  

For $K=2$, the improved geometric image-to-tikz ability of $M_1$ increases the number of reliable samples (including self-refined and transformation-augmented ones), driving a significant CLIP score gain for the subsequent model (denoted as $M_2$). At $K=3$, dataset growth slows: Despite sampling with replacement from the candidate geometric image pool, the model already achieves relatively mature geometric image-to-tikz capability, so the gain from newly added reliable samples in driving performance improvement diminishes. When $K=4$, dataset expansion diminishes further and the CLIP score of the model nearly converges. Hence, we set $K=4$ as the maximum iteration number for the experiments, balancing training cost and performance.

\begin{figure}[t]
  \centering
    \includegraphics[width=7.6cm, height=4.6cm]{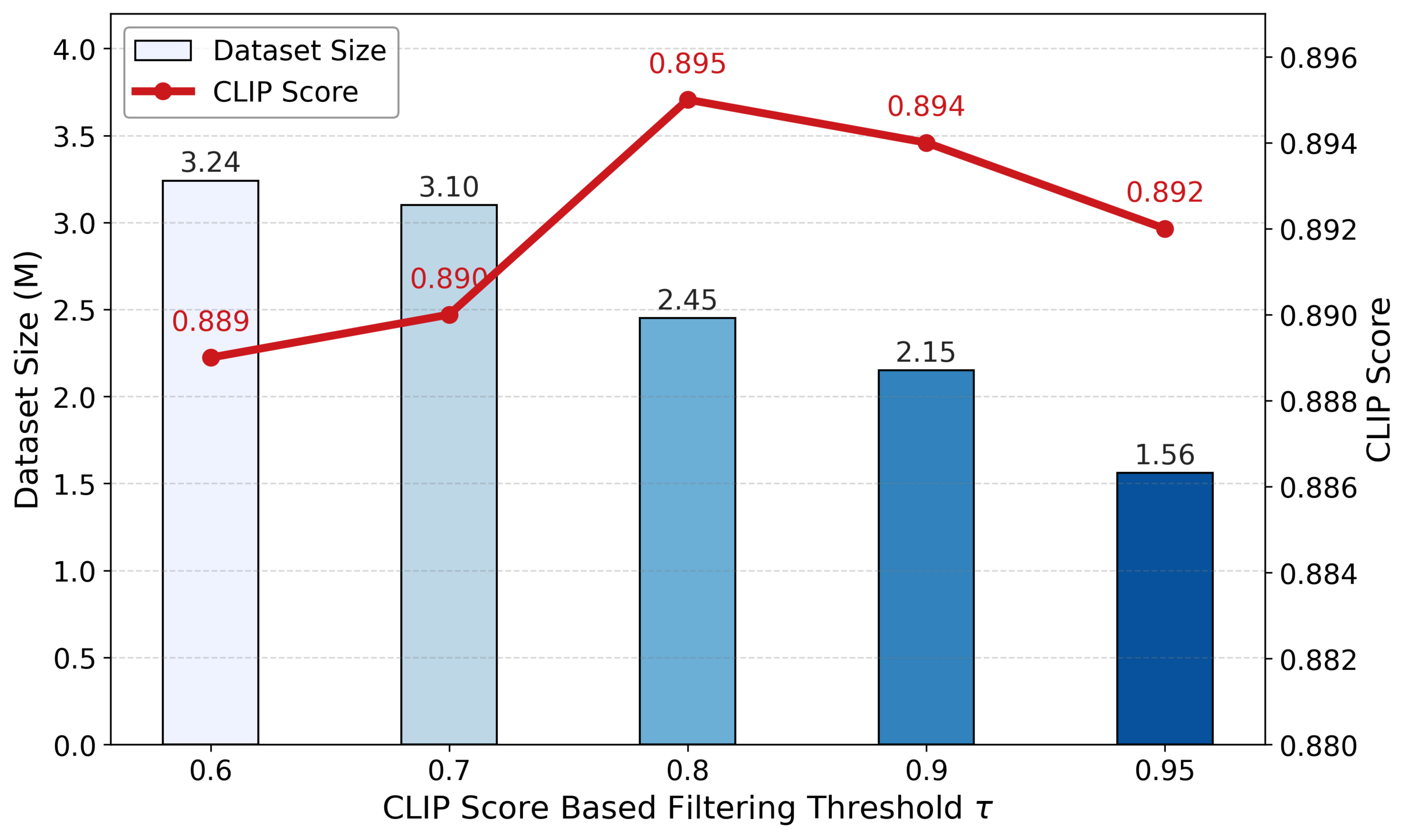}
    \caption{Ablation study of different values of CLIP score based filtering threshold $\tau$ for GeoTikzBridge-Base-8B, evaluated on the MathVista-GPS benchmark.}
    \label{fig: ablation_on_tau}
\end{figure}

\textbf{CLIP Score Based Filtering Threshold $\tau$.} We evaluate the effect of the CLIP score-based filtering threshold $\tau$ on both dataset scalability and model performance, Fig. \ref{fig: ablation_on_tau} presents the corresponding ablation study evaluated on the MathVista-GPS benchmark.  

For $\tau = 0.6$, the dataset size reaches 3.24M (the largest across all thresholds), while the CLIP score is 0.889. This indicates loose filtering introduces massive low-quality samples, limiting performance. At $\tau = 0.7$, the dataset size decreases slightly to 3.10M, and the CLIP score increases to 0.890. When $\tau = 0.8$, the dataset size decreases to 2.45M, yet the CLIP score peaks at 0.895, achieving the optimal trade-off between dataset scale and sample quality, where sufficient samples enable effective training, and high-quality samples drive the best performance. For $\tau = 0.9$ and $\tau = 0.95$, the dataset size further reduces to 2.15M and 1.56M, respectively, and the CLIP score decreases to 0.894 and 0.892, as insufficient training dataset restricts performance. Nevertheless, the CLIP scores achieved with the 2.15M and 1.56M datasets (at $\tau = 0.9$ and $\tau = 0.95$) remain higher than those obtained with the larger 3.24M and 3.10M datasets (at $\tau = 0.6$ and $\tau = 0.7$), indicating that the quality of training data is more critical than its quantity for this task. Hence, we set $\tau = 0.8$ as the CLIP score based filtering threshold, balancing dataset size and sample quality.

\begin{figure}[t]
  \centering
  \includegraphics[width=0.7\linewidth]{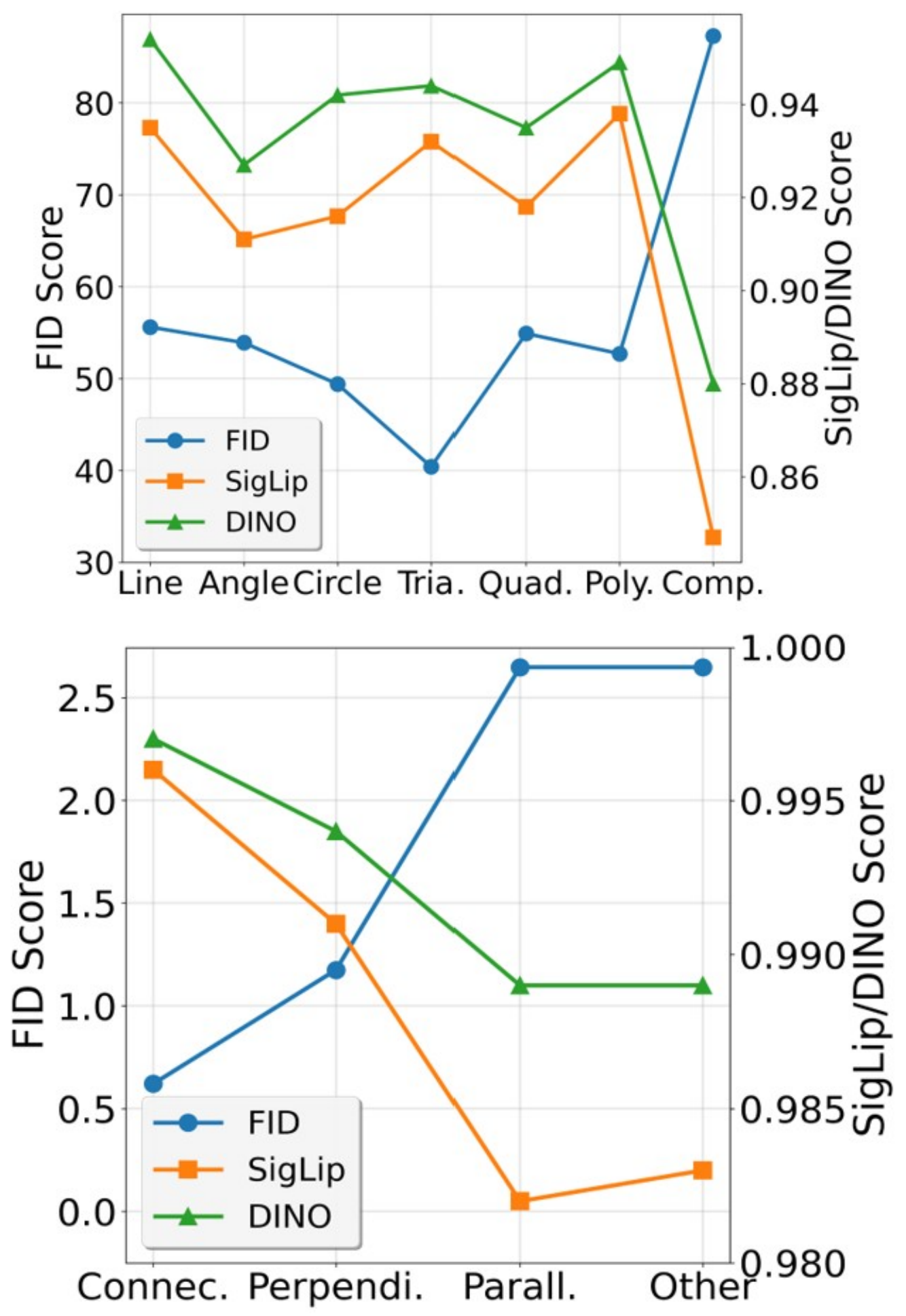}
   \caption{FID, SigLip/DINO scores of Base-8B model (up) and Instruct-8B model (bottom) across different geometric structures.}
   \label{fig: R1}
\end{figure}

\begin{figure}[t]
  \centering
  \includegraphics[width=0.8\linewidth]{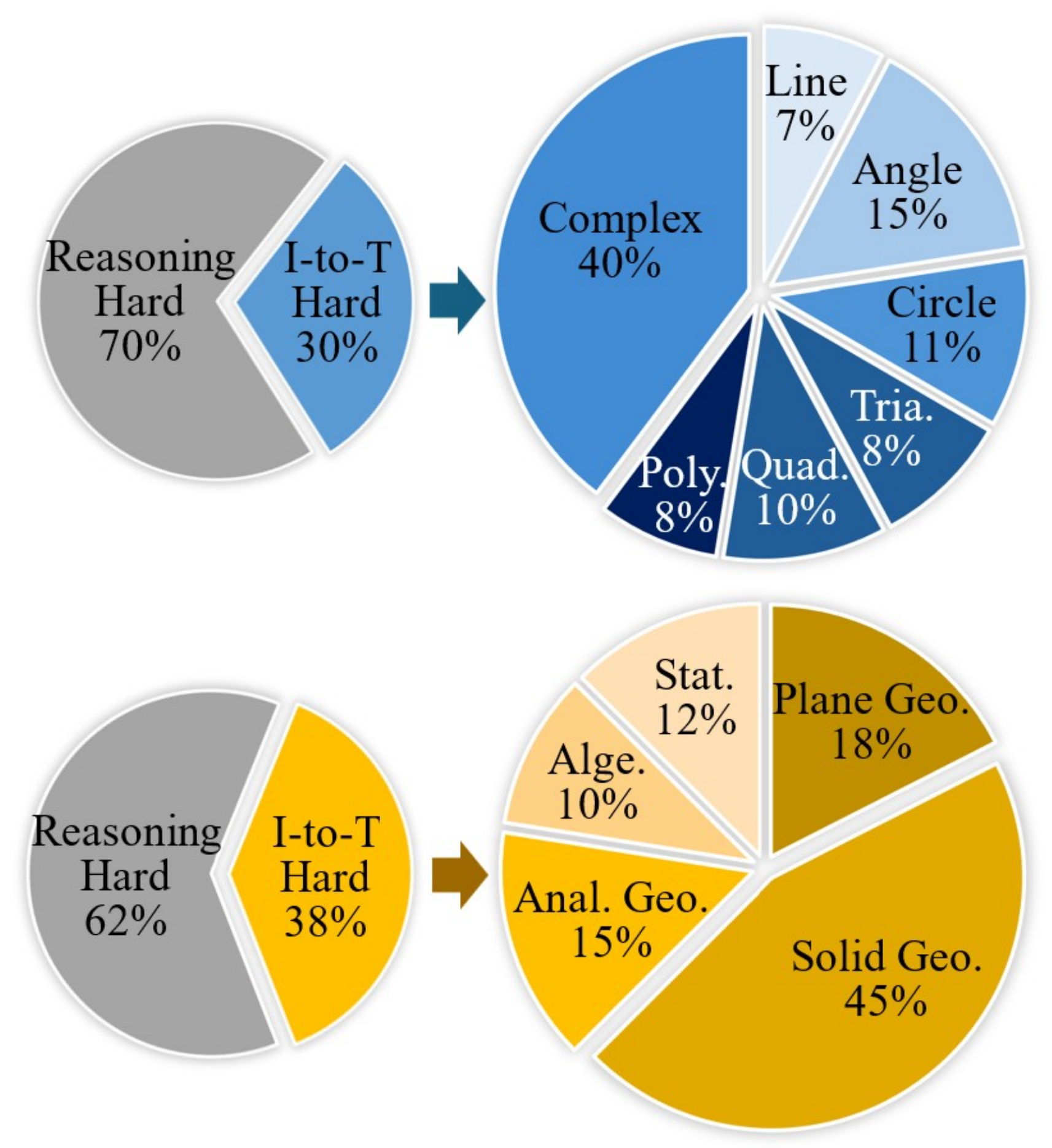}
   \caption{Average failure rate distributions of InternVL3.5-38B+GeoTikzBridge-Base-8B across different geometric structures (up) and mathematical domains (bottom).}
   \label{fig: R2}
\end{figure}

\begin{figure}[t]
  \centering
  \includegraphics[width=1.0\linewidth]{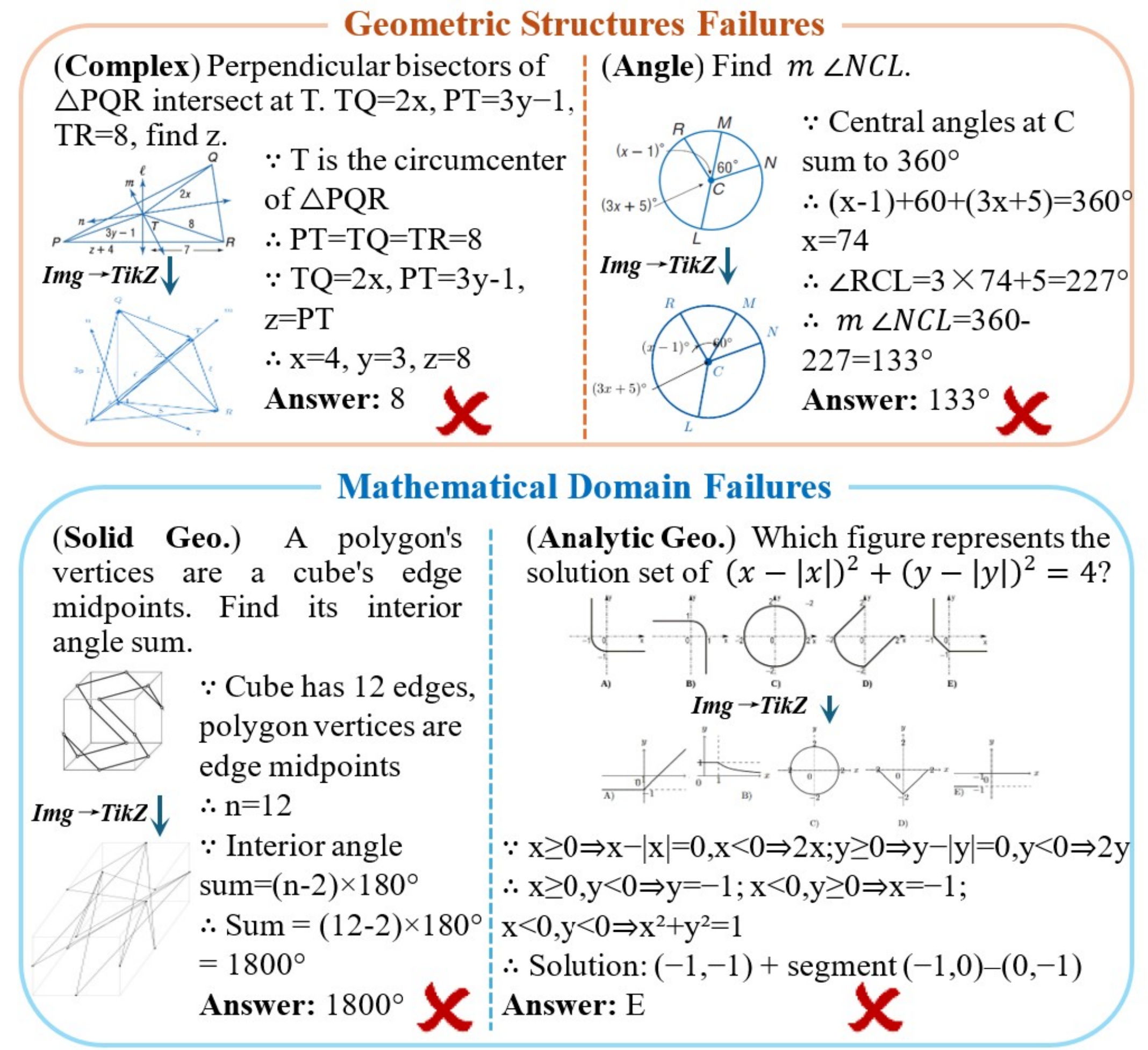}
   \caption{Failure cases of InternVL3.5-38B with tikz code input.}
   \label{fig: R3}
\end{figure}

\section{Failure Analysis} 
\label{sec: supple-F}

We analyze the failure cases of our proposed models on both geometric structure understanding and mathematical reasoning tasks. We first evaluate the GeoTikzBridge-Base-8B and GeoTikzBridge-Instruct-8B models across various geometric structures on the MathVista-GPS and GeoTikz-Instruct test sets, with results shown in Fig. \ref{fig: R1}. As shown from this figure, the GeoTikzBridge-Base-8B model suffers consistent performance degradation on complex figures, as overlong code tokens were truncated during training. The GeoTikzBridge-Instruct-8B model slightly underperforms on parallel line drawing tasks, due to the strict directional consistency constraint that raises higher requirements for generation precision.

Furthermore, we analyze problem-solving failure cases across geometric structures and mathematical domains, as shown in Fig. \ref{fig: R2}. I-to-T hard cases denote failure cases with an average score$\leq$0.9 across CLIP, SigLip and DINO metrics. The rest are reasoning hard cases. As shown from this figure, InternVL3.5-38B (with tikz code input) underperforms on tasks involving complex/angular geometric structures and solid/analytic geometry problems (examples in Fig. \ref{fig: R3}). These limitations stem from task difficulty, which we will address by developing an end-to-end reasoning paradigm in future work.

\section{Metric Model-Agnostic Effectiveness}
\label{sec: supple-G}

Though we use the CLIP model for evaluation, the performance gains of our models do not depend on any specific metric model. We provide additional SigLip and DINO evaluation results in Tables \ref{table: R3} and \ref{table: R4}. As shown in these two tables, our GeoTikzBridge-Base-8B and GeoTikzBridge-Instruct-8B models consistently outperform the FigCodifier baseline across CLIP, SigLip and DINO metrics. They also maintain significant advantages over state-of-the-art baselines including Qwen2.5VL-72B and InternVL3.5-38B. These results demonstrate the metric model-agnostic effectiveness of our approach.

\begin{table}[t]
    \renewcommand{\arraystretch}{1.3}
        \centering
        \relsize{-1}
        \caption{Performance comparison of instructed code generation methods across CLIP, SigLip, and DINO model metrics.}
        \label{table: R3}
        \begin{tabular}{p{3.4cm}>{\centering\arraybackslash}p{0.9cm}>{\centering\arraybackslash}p{0.9cm}>{\centering\arraybackslash}p{0.9cm}}
            \hline
            Method & CLIP & SigLip & DINO \\
            \hline
            FigCodifier & 0.936 & 0.940 & 0.957 \\
            ours-Base-8B & 0.967 & 0.970 & 0.978 \\
            ours-Instruct-8B & \bf 0.992 & \bf 0.992 & \bf 0.995 \\
            \hline
        \end{tabular}
\end{table}

\begin{table}[t]
\renewcommand{\arraystretch}{1.3}
\relsize{-2}
  \caption{Performance comparison of image-to-tikz methods across CLIP, SigLip, and DINO model metrics.}
  \label{table: R4}
  \centering
  \begin{tabular}{p{1.8cm}>{\centering\arraybackslash}p{0.6cm}>{\centering\arraybackslash}p{0.6cm}>{\centering\arraybackslash}p{0.6cm}>{\centering\arraybackslash}p{0.6cm}>{\centering\arraybackslash}p{0.6cm}>{\centering\arraybackslash}p{0.6cm}}
    \toprule
    \multirow{2}{*}{Method} & \multicolumn{3}{c}{DaTikZ} & \multicolumn{3}{c}{MathVista-GPS} \\
    \cmidrule{2-7} & ClIP &SigLip &DINO & ClIP &SigLip &DINO \\
    \midrule
    Qwen2.5-VL-72B & 0.795 & 0.803 & 0.832 & 0.858 & 0.880 & 0.911 \\
    Qwen3-VL-30B & 0.672 & 0.696 & 0.758 & 0.802 & 0.839 & 0.893 \\
    InternVL3.5-38B & 0.733 & 0.752 & 0.786 & 0.786 & 0.793 & 0.855 \\
    GLM4.5V-106B & \underline{0.806} & \underline{0.812} & \underline{0.834} & 0.769 & 0.773 & 0.838 \\
    FigCodifier & 0.785 & 0.791 & 0.819 & 0.884 & 0.906 & 0.922 \\
    \midrule
    ours-Base-8B & 0.804 & 0.808 & 0.832 & \underline{0.895} & \underline{0.919} & \underline{0.937} \\
    ours-Base-38B & \bf 0.813 & \bf 0.819 & \bf 0.875 & \bf 0.915 & \bf 0.941 & \bf 0.947  \\
    \bottomrule
  \end{tabular}
\end{table}

\section{Comparison vs. Latest Closed-Source Models.}
\label{sec: supple-H}

We compare the image-to-tikz performance of our GeoTikzBridge-Base-8B model with GPT-5.0 on the MathVista-GPS benchmark, with results reported in Table \ref{table: R1}. As seen from this table, our model outperforms GPT-5.0 across all image-to-tikz metrics. Furthermore, we also provide the tikz codes generated by GeoTikzBridge-Base-8B to latest closed-source models for mathematical reasoning. The results on MathVista-GPS benchmark reported in Table \ref{table: R2} demonstrates that applying our approach to closed models also yields improvements.

\begin{table}[t]
\renewcommand{\arraystretch}{1.3}
\relsize{-1}
  \caption{Performance comparison of image-to-tikz methods with GPT-5.0 on the MathVista-GPS benchmark across five metrics.}
    \label{table: R1}
  \centering
  \begin{tabular}{p{1.8cm}>{\centering\arraybackslash}p{0.7cm}>{\centering\arraybackslash}p{0.7cm}>{\centering\arraybackslash}p{0.8cm}>{\centering\arraybackslash}p{0.9cm}>{\centering\arraybackslash}p{0.9cm}
  }
    \toprule
    Method & FID$\downarrow$  & CSR$\uparrow$ & CLIP$\uparrow$ & SigLip$\uparrow$ & DINO$\uparrow$ \\
    \midrule
    GPT-5.0 & 44.8 & 96.9\% & 0.880 & 0.903 & 0.921\\
    ours-Base-8B & \bf36.0 & \bf 97.1\% & \bf 0.895 & \bf 0.919 & \bf 0.937\\
    \bottomrule
  \end{tabular}
\end{table}

\begin{table}[t]
\renewcommand{\arraystretch}{1.3}
\relsize{-1}
  \caption{Mathematical reasoning performance of latest closed-source models on the MathVista-GPS benchmark. Scores are reported as original model / model + GeoTikzBridge-Base-8B.}
  \label{table: R2}
  \centering
  \begin{tabular}{p{3.4cm}>{\centering\arraybackslash}p{4.0cm}}
    \toprule
    Method & Accuracy (Base/+Ours) \\
    \midrule
    GPT-5.0 & 0.891/\bf 0.937  \\
    Claude-4.5-Sonnect & 0.774/\bf 0.846  \\
    Gemini-3-Pro & 0.779/\bf 0.923  \\
    \bottomrule
  \end{tabular}
\end{table}

\section{Supplementary Visualization}
\label{sec: supple-I}

\begin{figure*}[t]
  \centering
    \includegraphics[width=17.5cm, height=20.5cm]{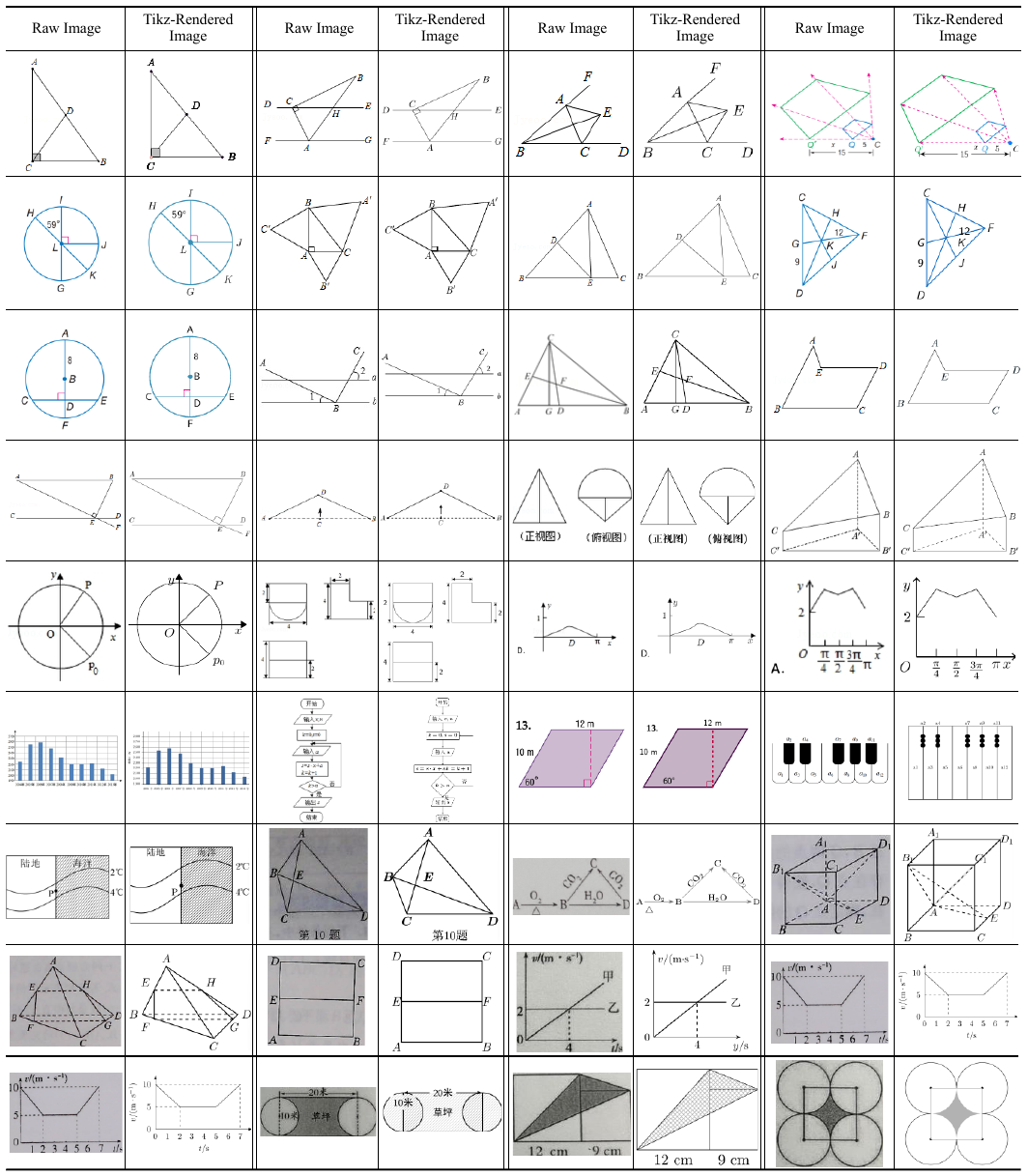}
    \caption{Visualization results of tikz codes generated by GeoTikzBridge-Base, consisting of input images (either geometric or non-geometric) and the corresponding figures rendered from our predicted tikz codes.}
    \label{fig: GeoTikzBridge-Base_results_more}
\end{figure*}

\begin{figure*}[t]
  \centering
    \includegraphics[width=17.5cm, height=20.5cm]{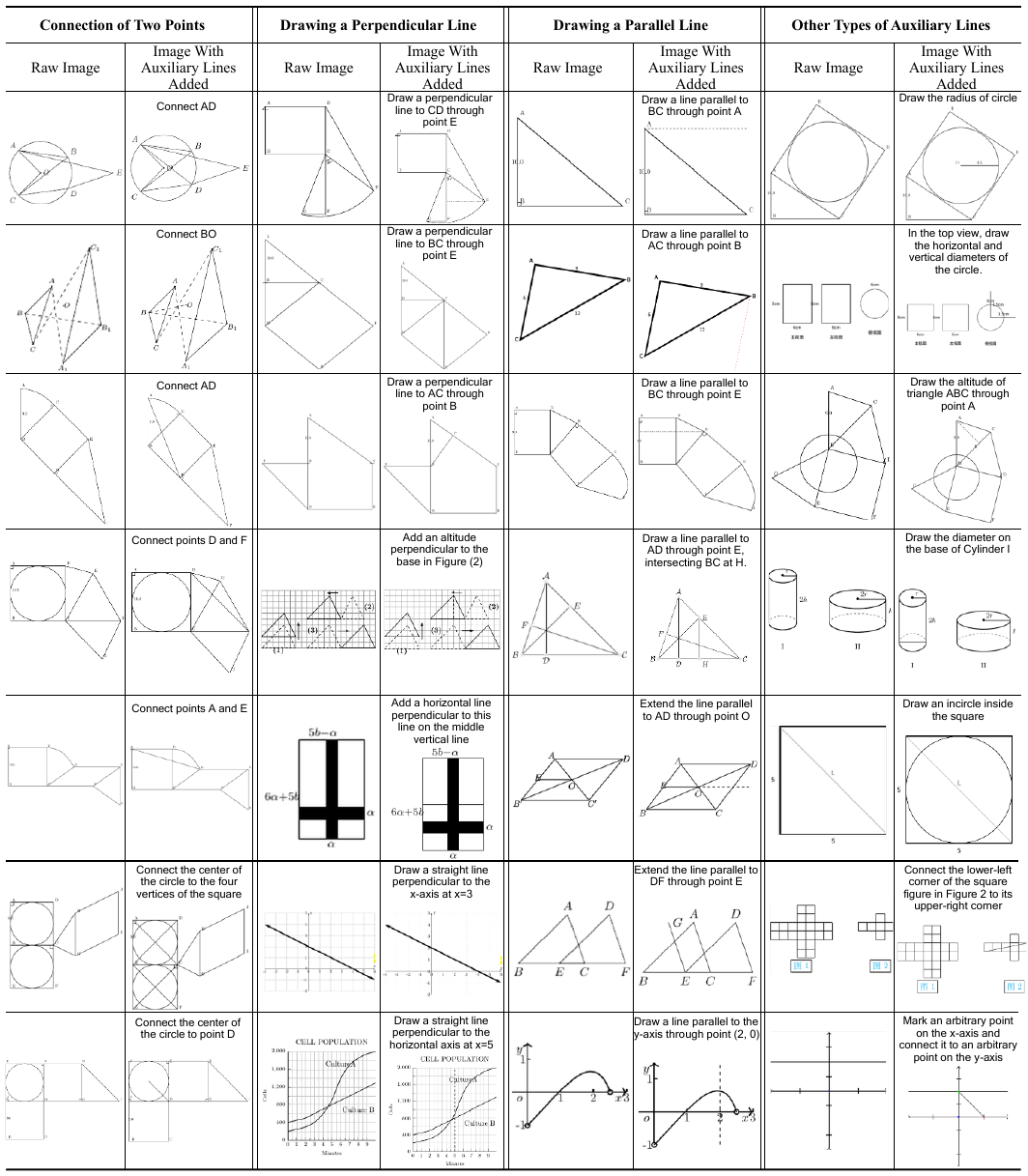}
    \caption{Visualization results of instruction-driven auxiliary line generation, including the input geometric images, the natural language instructions for auxiliary line addition, and the rendered figures from the generated tikz codes.}
    \label{fig: aux_line_results_more}
\end{figure*}

\begin{figure*}[t]
  \centering
    \includegraphics[width=17.5cm, height=20.5cm]{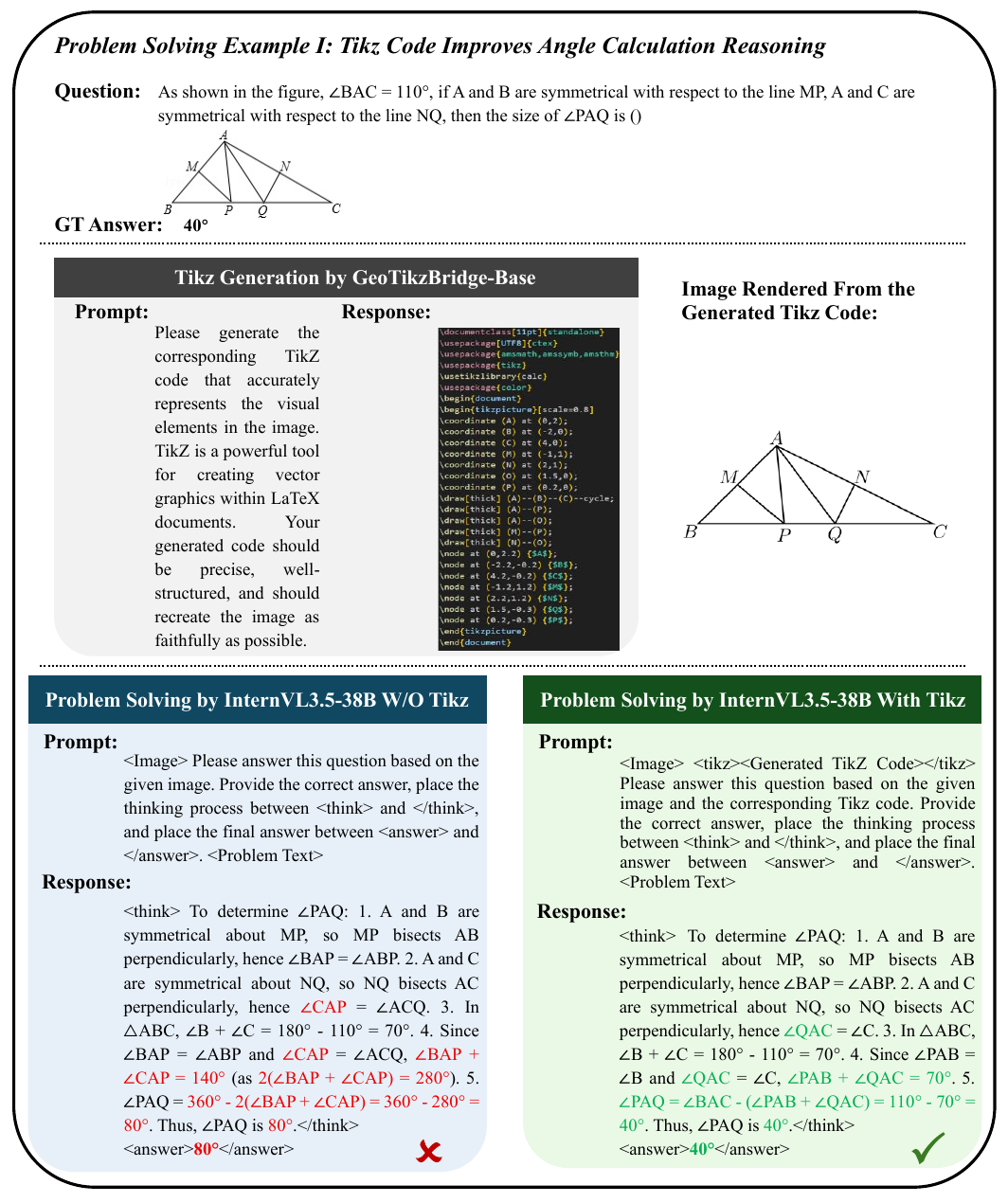}
    \caption{Example I: Angle calculation problem solving comparison with and without tikz code. Tikz clarifies geometric relationships to avoid perceptual errors in angle reasoning.}
    \label{fig: aux_line_results0}
\end{figure*}

\begin{figure*}[t]
  \centering
    \includegraphics[width=17.5cm, height=20.5cm]{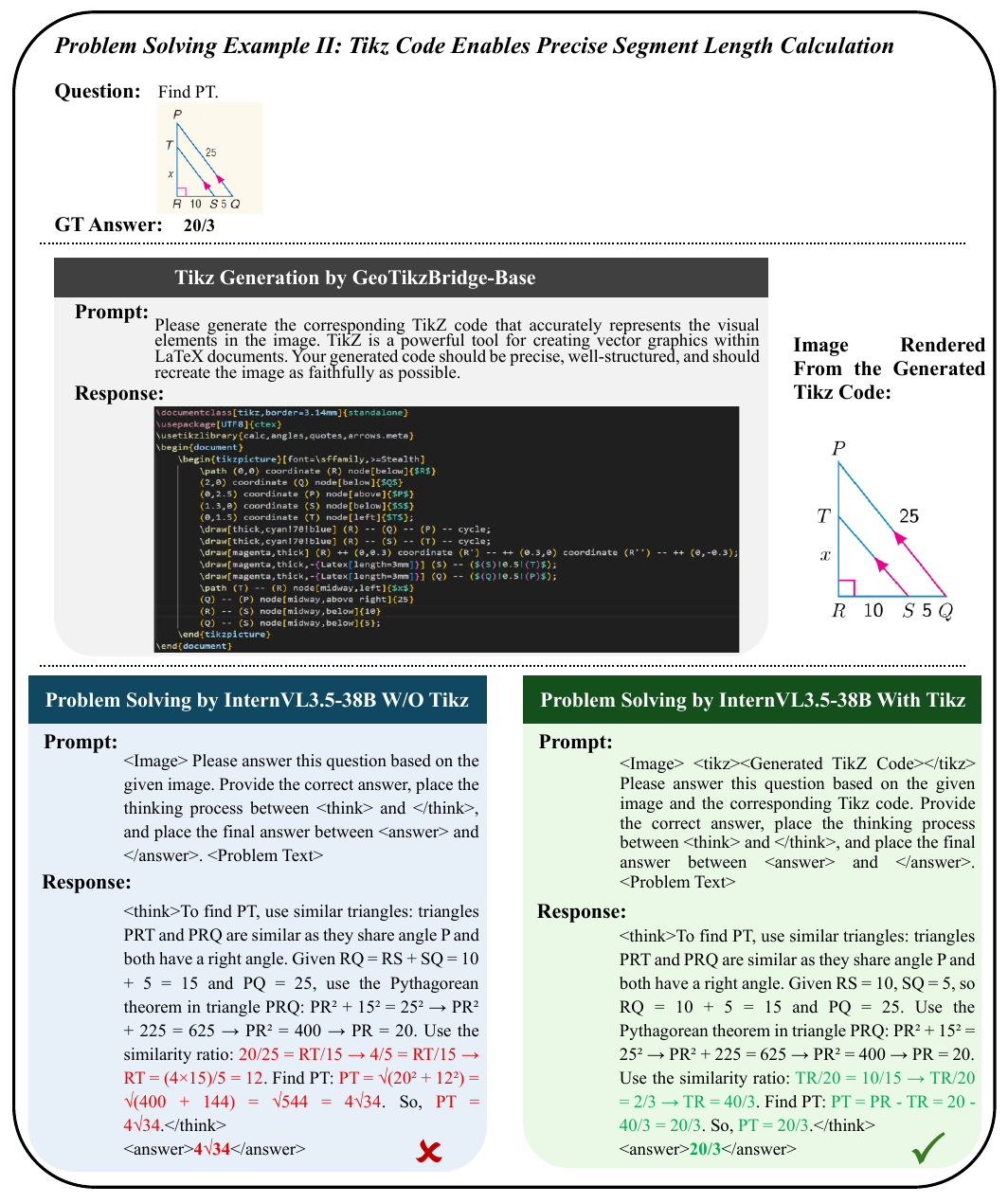}
    \caption{Example II: Segment length problem solving comparison with and without tikz code. Tikz facilitates accurate proportion and theorem application for length reasoning.}
    \label{fig: aux_line_results0_another}
\end{figure*}

\begin{figure*}[t]
  \centering
    \includegraphics[width=17.5cm, height=20.5cm]{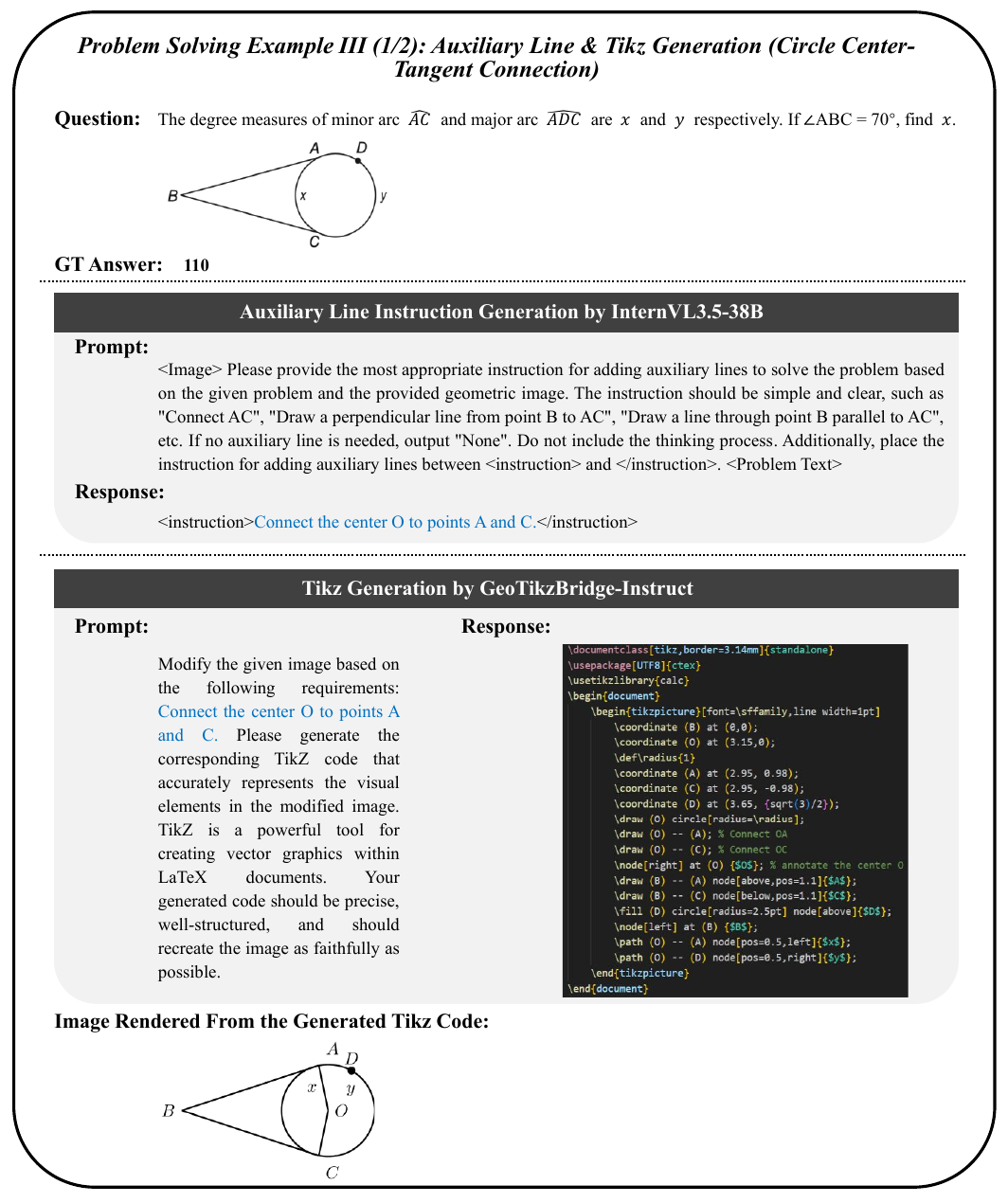}
    \caption{Example III (1/2): Auxiliary line instruction (connecting circle center and tangent points) and tikz code generation. The line manifests radius-tangent perpendicularity.}
    \label{fig: aux_line_results1_Page1}
\end{figure*}

\begin{figure*}[t]
  \centering
    \includegraphics[width=17.5cm, height=20.5cm]{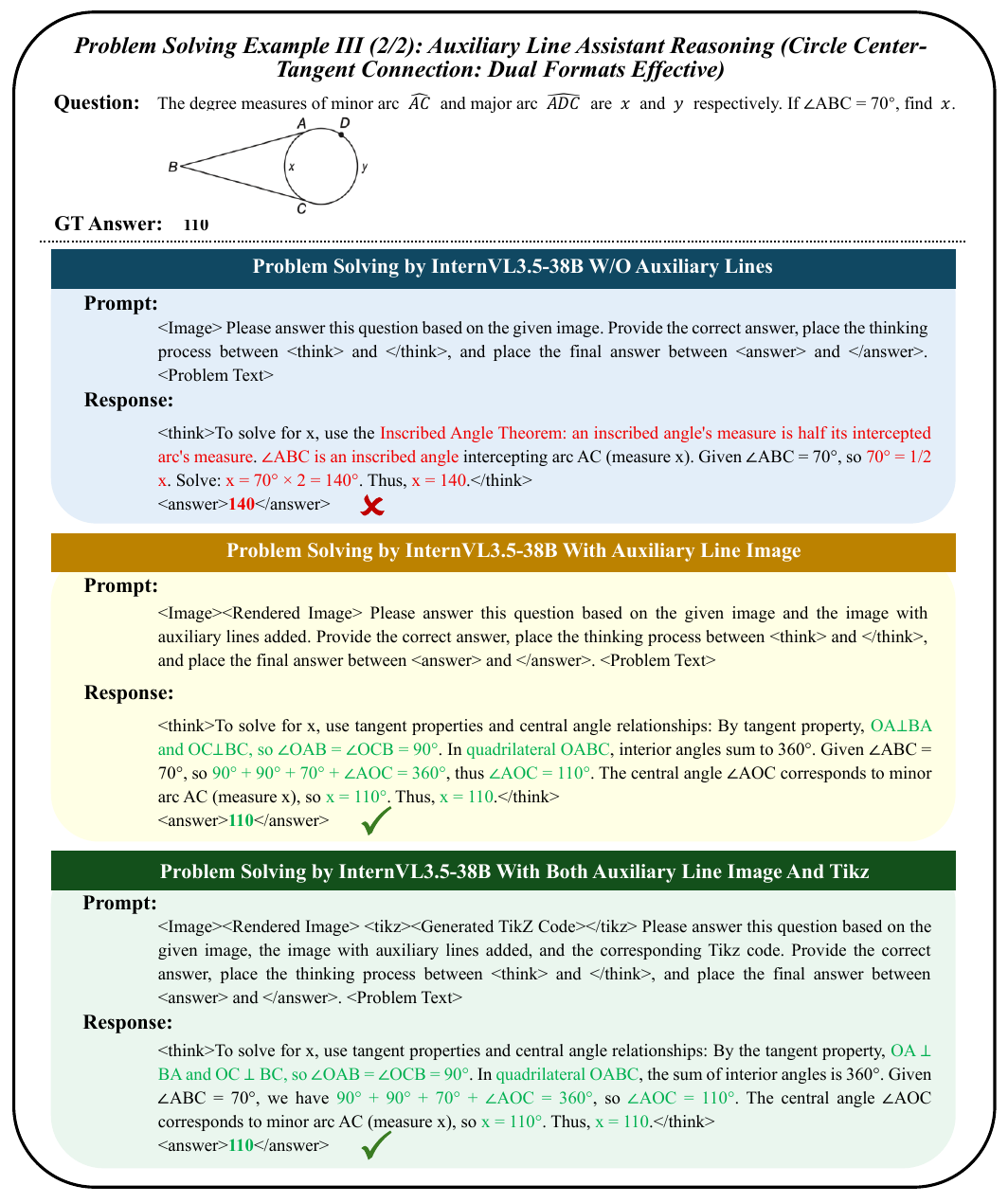}
    \caption{Example III (2/2): Problem solving comparison under different forms, with incorrect analysis/answers highlighted in red and key correct ones in green. Both auxiliary line image and its tikz combination enable correct reasoning via explicit perpendicularity recognition.}
    \label{fig: aux_line_results1_Page2}
\end{figure*}

\begin{figure*}[t]
  \centering
    \includegraphics[width=17.5cm, height=20.5cm]{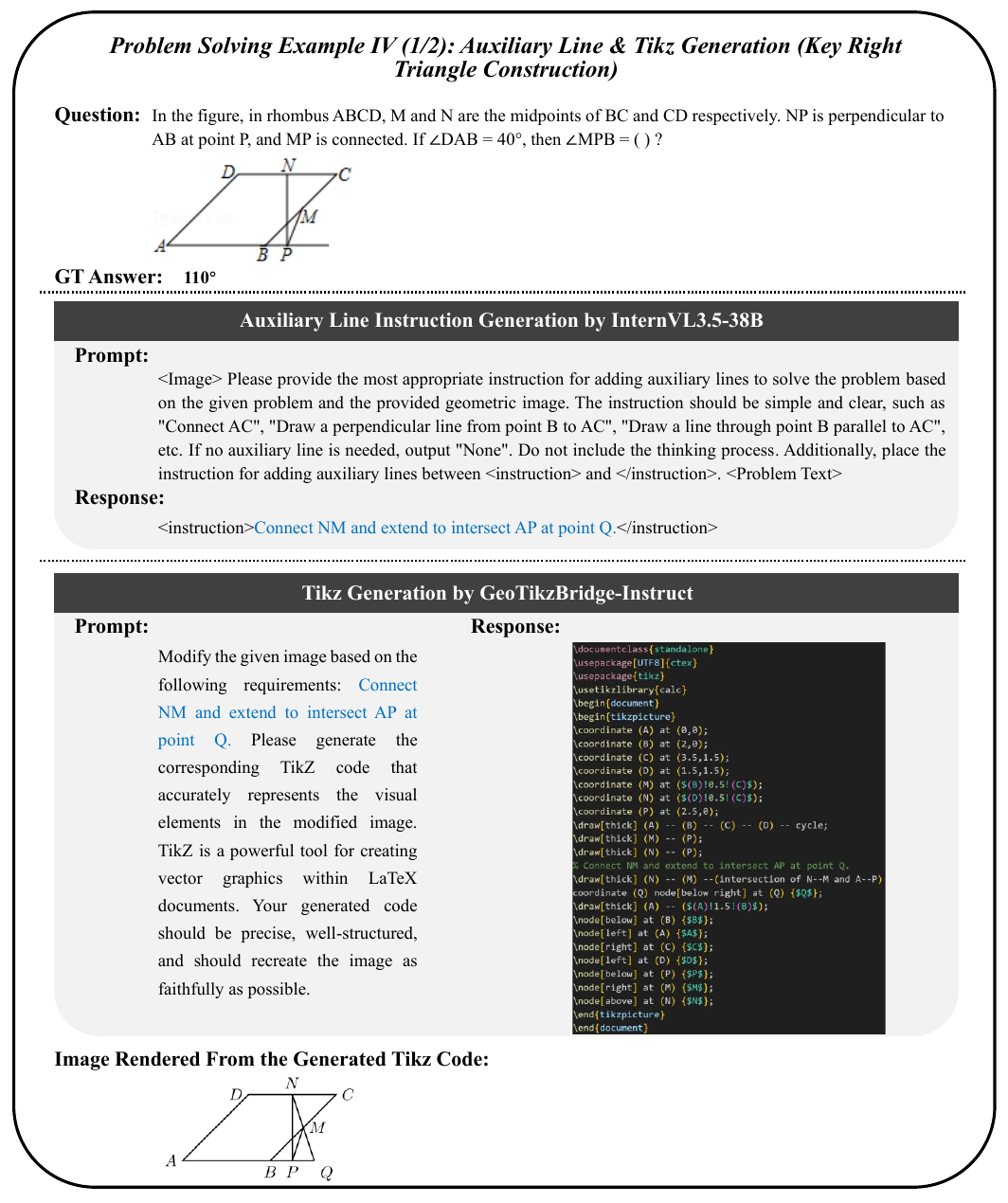}
    \caption{Example IV (1/2): Auxiliary line instruction (extending midpoint connection) and tikz code generation. The line constructs a key right triangle by forming intersection $Q$.}
    \label{fig: aux_line_results2_Page1}
\end{figure*}

\begin{figure*}[t]
  \centering
    \includegraphics[width=17.5cm, height=20.5cm]{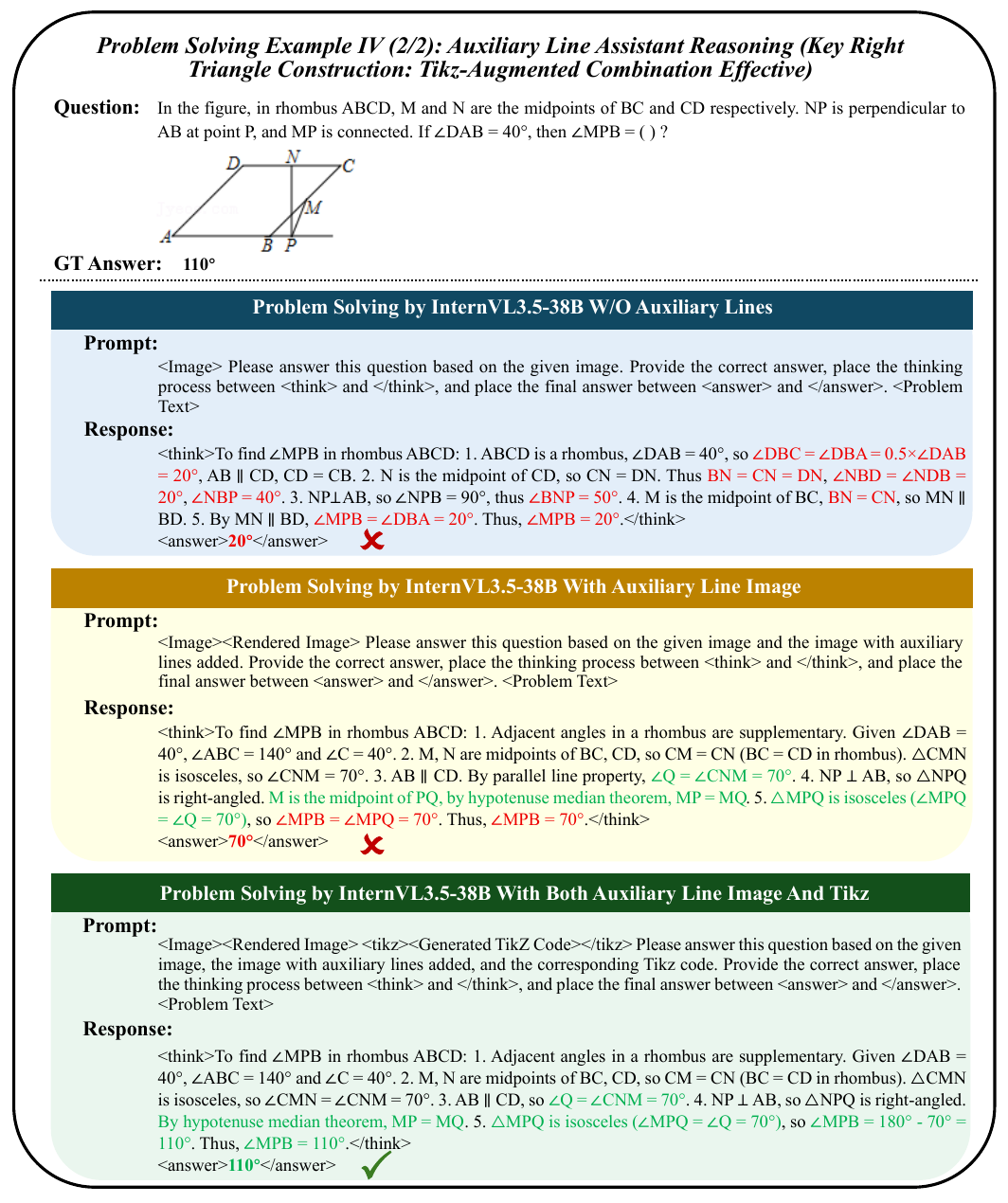}
    \caption{Example IV (2/2): Problem solving comparison under different forms, with incorrect analysis/answers highlighted in red and key correct ones in green. The auxiliary line image provides a visual foundation for constructing the right triangle $\triangle$NPQ. Its combination with tikz code achieves correct reasoning, since tikz explicitly encodes the spatial structure of the constructed right triangle.}
    \label{fig: aux_line_results2_Page2}
\end{figure*}

To supplement the results partially presented in the main text, we provide additional examples for the image-to-tikz task and the instruction-based image-to-tikz task in Fig. \ref{fig: GeoTikzBridge-Base_results_more} and Fig. \ref{fig: aux_line_results_more}, respectively. As shown in Fig. \ref{fig: GeoTikzBridge-Base_results_more}, our model still maintains precise capture of key geometric components (line segments, angles, closed polygons, and semantic labels) in these additional examples. Fig. \ref{fig: aux_line_results_more} further verifies that our model generates auxiliary lines more compliant with given instructions while preserving the characteristics of the original geometric figures.

As a complement to the quantitative results of both tikz assistant problem-solving task (Example I in Fig. \ref{fig: aux_line_results0} and Example II in Fig. \ref{fig: aux_line_results0_another}) and auxiliary line assistant problem-solving task (Example III in Fig. \ref{fig: aux_line_results1_Page1} and Fig. \ref{fig: aux_line_results1_Page2}, as well as Example IV in Fig. \ref{fig: aux_line_results2_Page1} and Fig. \ref{fig: aux_line_results2_Page2}) presented in Table 2 and Table 4 of the main text (Sec. 4.2), we provide four representative qualitative examples. Consistent with the main text's evaluation setup: For the first two examples, InternVL3.5-38B is used for geometric reasoning; For the last two examples, InternVL3.5-38B first predicts auxiliary line instructions for original geometric problems, these instructions and raw images are then fed into our instruction-based model GeoTikzBridge-Instruct to generate tikz codes with auxiliary lines added, and the generated tikz codes and/or their rendered images are finally input to InternVL3.5-38B for geometric reasoning.

Notably, the examples align with key observations from Table 2 and Table 4: Examples I and II demonstrate that the model corrects its image perception errors after incorporating tikz codes; Example III yields an incorrect solution without auxiliary lines but correct ones with either augmented image or tikz code, demonstrating that introducing auxiliary lines, either in the form of augmented images or augmented images combined with tikz codes, benefit geometric problem solving; Example IV fails in both models ``W/O Auxiliary Lines" and ``With Auxiliary Line Image", with only ``With Both Auxiliary Line Image And Tikz" leading to a correct solution, further demonstrating that tikz codes outperforms augmented images in facilitating precise geometric reasoning.

\end{document}